\documentclass[11pt]{article}

\usepackage[preprint]{acl}
\usepackage[normalem]{ulem}
\usepackage{times}
\usepackage{latexsym}
\usepackage[T1]{fontenc}

\usepackage[utf8]{inputenc}

\usepackage{microtype}

\usepackage{inconsolata}

\usepackage{graphicx}
\usepackage{booktabs}
\usepackage{tabularx}
\usepackage{makecell}
\usepackage{array}
\usepackage{amsmath}
\usepackage{microtype}
\usepackage{xcolor}
\usepackage{alltt}
\usepackage{algorithm}
\usepackage{algorithmic}
\definecolor{eoiGreen}{RGB}{36,135,67}
\definecolor{baseRed}{RGB}{190,54,54}

\newcolumntype{Y}{>{\raggedright\arraybackslash}X}

\definecolor{AcademicRed}{RGB}{180, 40, 40}
\definecolor{AcademicGreen}{RGB}{40, 140, 80}

\title{Clearing the Fog: Towards Installing and Refining Proactive Exploration Capabilities in LLM Agents}

\newcommand{\ours}{\textsc{SaFaRi}\xspace}

\author{
Zhizhao Guan$^{1,4}$
\quad  Chen Huang$^{2}$\thanks{Corresponding author.} 
\quad Ziming Liu$^{1}$
\quad \textbf{Hongru Liang}$^{1,4}$ 
\quad \textbf{Wenqiang Lei}$^{1,4}$  \\
\quad \textbf{See-Kiong Ng}$^{2}$
\quad \textbf{Tat-Seng Chua}$^{2}$
\quad \textbf{Anthony G Cohn}$^{3}$\\
\\[-1em]
$^{1}$ Sichuan University  
\quad $^{2}$ National University of Singapore  \quad
$^{3}$ University of Leeds \\
$^{4}$Engineering Research Center of Machine Learning and Industry Intelligence \\
\texttt{guanzhizhao@stu.scu.edu.cn, huang\_chen@nus.edu.sg}
}

\begin{document}
\maketitle
\begin{abstract}
We study proactive exploration in LLM agents, i.e., the ability to explore an environment to acquire information that improves future decision-making. In this regard, we first identify two fundamental bottlenecks that hinder this capability and then propose \ours, a novel method designed to instill and refine proactive exploration. Specifically, \ours\ consists of two components: (1) Exploratory Data Construction, which synthesizes exploration-rich trajectories to mitigate the hindsight bias of standard demonstrations; and (2) RL Optimization with Contrastive Signal Guidance, which leverages contrastive trajectory pairs to distinguish productive exploration from redundant wandering. Extensive experiments demonstrate the effectiveness of \ours\ and provide insights into the characteristics of proactive exploration. Our code is available at: \url{https://github.com/GuanZhizhao/SAFARI}.
\end{abstract}

\section{Introduction}

\begin{figure}[th] 
    \centering
    \includegraphics[width=\columnwidth]{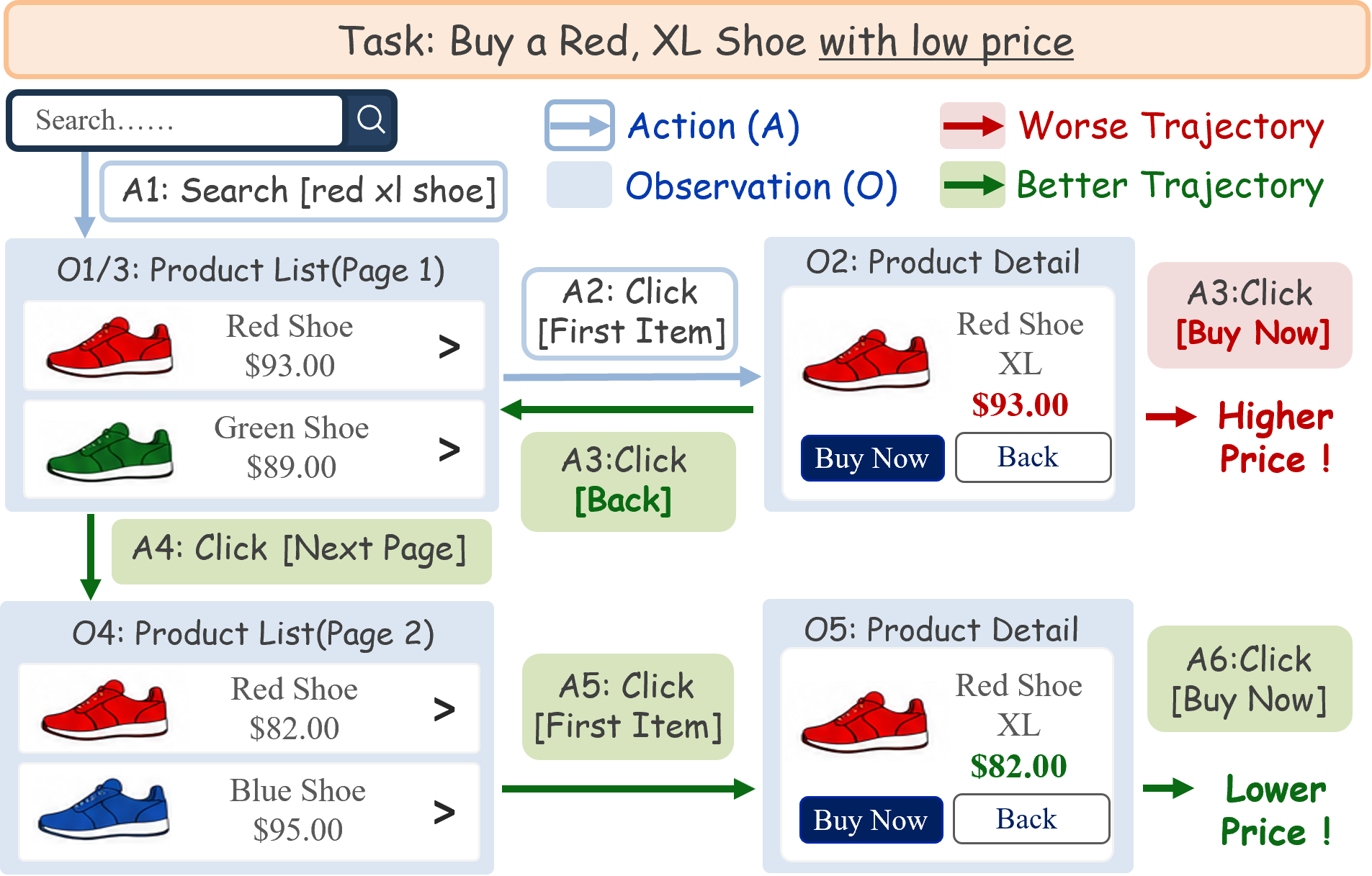} 
    \caption{ \textbf{Proactive exploration in online shopping.} Agent proactively executes strategic backtracking and pagination to locate the product with better price.}
    \vspace{-5.5mm}
    \label{fig:proactive_example}
\end{figure}

In multi-turn interactive tasks like online shopping \cite{yao2022webshop,zhou2023webarena,koh2024visualwebarena}, agents must continuously make decisions based on information acquired during interaction. Taking Figure \ref{fig:proactive_example} for example, when purchasing a low-cost shoe with specific attributes, the agent must decide whether to settle for a currently satisfactory item (\textcolor{AcademicRed}{$A_3$}), or continue browsing subsequent pages for a potentially better alternative (\textcolor{AcademicGreen}{$A_6$}). Such scenarios require a critical capability: \textbf{Proactive Exploration}, which is the ability to autonomously expand its action trajectory beyond immediately local optima. This capability is fundamental for real-world agents, where optimal decisions often cannot be derived from the current observation alone, but instead emerge through strategic exploration over extended interaction horizons \cite{fang2025information}.

As Large Language Models (LLMs) continue to advance, LLM-based agents have excelled in multi-turn interactive tasks \cite{yao2022webshop,zhou2023webarena}. The prevailing paradigm for developing these agents involves Supervised Fine-Tuning (SFT) on expert trajectories \cite{lin2023swiftsage,rita2024countering,chen2023fireact,yin2023lumos}, followed by Reinforcement Learning (RL) to iteratively refine decision-making \cite{carta2023grounding,tan2024true,song2024trial,xiong2024watch,yu2024exact}. However, our preliminary experiments (Section~\ref{sec:exploration_action_distrubution_analysis}) reveal that existing agents severely lack the capability for proactive exploration. In scenarios like online shopping, they exhibit rigid behavioral patterns, rarely executing exploratory actions (e.g., navigating to the "Next Page"). Consequently, they fail to strategize over extended horizons, frequently settling prematurely for suboptimal choices and yielding inferior overall rewards. We attribute this failure to a two-stage bottleneck inherent in the standard SFT-RL pipeline:
\textbf{1) Hindsight Bias in SFT Data}: Expert trajectories inherently strip away essential exploratory steps. Standard SFT merely conditions agents for reactive execution, failing to instill proactive exploration capabilities.
\textbf{2) Exploration Collapse during RL}: Because SFT-ed agents naturally lack exploratory tendencies, they rarely sample such behaviors during RL rollouts, making it nearly impossible to discover their long-term benefits. Compounded by a lack of guidance, the RL process inevitably fails to reinforce proactive exploration.

To this end, we propose \ours, a novel method designed to in\uline{{S}}till
\uline{{A}}nd 
re\uline{{F}}ine
pro\uline{{A}}ctive 
explo\uline{{R}}at\uline{{I}}on capabilities in LLM agents. Specifically, it consists of two modules: \textbf{1) Exploratory Data Construction}. Instead of human expert trajectories, we synthesize exploration-rich data directly from a powerful LLM expert. By employing a tree-structured context modeling, we explicitly distill diverse exploratory paths from the LLM expert, coupled with a customized screening mechanism to curate high-quality, exploration-oriented trajectories for the student agent's SFT initialization. \textbf{2) RL with Contrastive Signal Guidance.} We construct contrastive preference pairs by pitting a student-generated candidate action against the reference action at identical interaction states. By estimating their expected future returns via Monte Carlo (MC) rollouts, these contrastive signals provide fine-grained supervision to optimize the agent's decision boundary. 
Thus \ours\ alternates between strategic exploration and direct task execution,  improving the task performance.


We conduct extensive experiments across diverse benchmarks to evaluate \ours. Results demonstrate that \ours\ significantly outperforms competitive baselines in both task success ($\sim$10\%-15\%, on average) and exploration efficiency ($\sim$8\%-18\%). 
Additionally, our analyses illustrate that true performance improvements stem from exploration efficiency. \ours\ learns to acquire task-relevant information while avoiding redundant interactions, whereas existing methods remain either passive or inefficient. Moreover, exploration benefits also grow with task difficulty. \ours\ automatically scales its exploratory behavior in harder environments and achieves the largest gains on tasks that require substantial information gathering before decision-making.
Our contributions are as follows:
\begin{itemize}[leftmargin=3mm,itemindent=0.05cm, itemsep=0.5pt]
    \item We highlight the critical necessity of proactive exploration for LLM-based agents and identify two fundamental bottlenecks in existing methods.
    \item We propose \ours, designed to instill and refine proactive exploration capabilities in LLM agents. It features two customized modules to directly mitigate the two bottlenecks, respectively.
    \item We conduct extensive experiments across various benchmarks to demonstrate the effectiveness and characteristics of \ours.
\end{itemize}

\section{Related Work}
Early interactive agents primarily relied on prompt-based planning \cite{madaan2023self,yao2023react} or imitation learning \cite{lin2023swiftsage,chen2023fireact}, limiting their behavior to observed state--action distributions \cite{seo2024mitigating}. To improve adaptability, recent work has introduced exploration-driven RL training and iterative self-improvement \cite{song2024trial,xiong2024watch,yu2024exact}. However, hindsight bias and exploration collapse often result in insufficient proactive exploration and yielding suboptimal solutions. To address this issue, recent research has proposed error recovery after explicit failures, achieved by corrective trajectories \cite{wang2025steca, chen2025atlas} and MCTS-based self-improvement \cite{yuan2025agent, yu2025exact}. Nevertheless, since these methods intervene only upon failure, they discourage exploration beyond reward-sufficient trajectories, often converging to local optima.

\begin{table*}[t]
    \centering
    \small
    \renewcommand{\arraystretch}{1.15} 
    \setlength{\tabcolsep}{5pt} 
    \resizebox{0.9\textwidth}{!}{%
    \begin{tabular}{@{} l c @{\hspace{1.5em}} cccc @{\hspace{1.5em}} cccc @{}}
        \toprule
        \multirow{2}{*}{\textbf{Method}} & \multirow{2}{*}{\textbf{Avg. Len}} & \multicolumn{4}{c}{\textbf{Task-Oriented Actions (\%)}} & \multicolumn{4}{c}{\textbf{Exploratory Actions (\%)}} \\
        \cmidrule(lr){3-6} \cmidrule(l){7-10}
        & & Search & Open & Select & Buy & Next & Back & Home & Total \\
        \midrule
        SFT-Only \cite{chen2023fireact} & 3.33 & 30.15 & 30.15 & 9.85 & 29.85 & 0.00 & 0.00 & 0.00 & 0.00 \\
        ETO \cite{song2024trial}& 4.32 & 23.13 & 22.99 & 28.69 & 24.84 & 0.12 & 0.23 & 0.00 & 0.35 \\
        IPR \cite{xiong2024watch}& 4.39 & 23.22 & 22.18 & 29.89 & 24.60 & 0.00 & 0.11 & 0.00 & 0.11 \\
        STeCa \cite{wang2025steca}& 4.87 & 20.85 & 24.59 & 28.22 & 25.31 & 0.00 & 0.21 & 0.83 & 1.04 \\ \midrule
        Human Expert Trajectories \cite{yao2022webshop} & 3.64 & 28.20 & 27.69 & 16.51 & 25.31 & 0.00 & 0.10 & 0.00 & 0.00 \\
        \bottomrule
    \end{tabular}
    }
    \setlength{\abovecaptionskip}{2pt}
\setlength{\belowcaptionskip}{0pt}
    \caption{Comparison of action across baselines and human expert SFT trajectories in WebShop. }
    \vspace{-6mm}
    \label{tab:action_distribution}
\end{table*}

Another line of work improves decision-making via structured representations. Tree-based reasoning methods enhance state visibility and search efficiency \cite{yao2023tree, zhou2023language, koh2024tree}. Recent methods construct interaction graphs \cite{gandhi2025go} or exploit webpage hierarchies \cite{zhang2026webnavigator} to guide navigation. However, these methods are designed for web environments and operate as external search tools rather than policy-level mechanisms, limiting their applicability to broader domains such as database manipulation and scientific environments.

\section{Preliminary Experiments}
\label{sec:exploration_action_distrubution_analysis}
We investigate the exploration capabilities of existing methods from two aspects: agents' behavioral action distributions and their training dataset.


\subsection{Evaluation Setup}
\noindent\textbf{Dataset \& Exploratory Actions}. 
We select WebShop \cite{yao2022webshop} as our testbed due to its inherent decoupling of exploratory behaviors from terminal task execution, allowing for a precise analysis of the agent's exploratory tendencies. Specifically, we categorize the action space into: (1) \uline{Task-Oriented Actions} (\textit{Search}, \textit{Open}, \textit{Select}, \textit{Buy}), dedicated to immediate execution; and (2) \uline{Exploratory Actions} (\textit{Next}, \textit{Back}, \textit{Home}) designed to actively explore the environment. Ultimately, we conduct a statistical analysis of these action distributions to diagnose the cause of agent failure.

\noindent\textbf{Baselines.} 
We evaluate three categories of baselines: 
1) \uline{SFT-Only} \cite{chen2023fireact}, where agents are fine-tuned exclusively on standard expert trajectories; 2) \uline{SFT-RL}, the prevailing paradigm that sequentially applies SFT and RL, represented here by two state-of-the-art approaches, \textit{ETO} \cite{song2024trial} and \textit{IPR} \cite{xiong2024watch}, which employ trajectory-level and step-level supervision, respectively; and 3) \uline{SFT-RL-Correction}, represented by \uline{STeCa} \cite{wang2025steca}, which augments the standard SFT-RL with post-hoc corrections by advanced LLM. All baselines utilize Meta-Llama-3.1-8B-Instruct~\cite{grattafiori2024llama}.

\subsection{Main Findings}
Table~\ref{tab:action_distribution} summarizes the results of the action distribution analysis. Evidently, both existing methods and the underlying SFT data profoundly lack exploratory actions. We highlight two bottlenecks. \noindent\uline{1) Hindsight Bias in SFT Data}. Expert trajectories strip away essential exploratory steps, with exploratory actions comprising a negligible 0.10\%. Consequently, initializing agents on such biased corpora merely conditions them for purely reactive execution. This is exemplified by the SFT-Only baseline on WebShop, which exhibits near-zero exploratory actions (e.g., \textit{Next} or \textit{Back}), failing to instill the core capabilities required for proactive exploration. \uline{2) Exploration Collapse during RL}. Even with RL (ETO, IPR) or post-hoc correction (STeCa), we observe no qualitative shift in exploratory action distributions. Instead, these methods converge toward myopic optimization, merely refining item \textit{selection} (i.e., $\sim$20\% gains in distribution) within the current observation. Lacking exploratory tendencies in the SFT-ed policy, exploratory actions are rarely sampled during rollouts, preventing the agent from discovering their long-term rewards. As corroborated by our case studies (Appendix~\ref{app:case_study}) and main experiments (Section~\ref{sec:main_result}), this deficit leads to significant performance degradation in complex tasks. Thus, it is imperative to incorporate an exploration-rich SFT dataset and RL mechanisms that shift the agent's behavior from reactive execution to proactive exploration.

\begin{figure*}[ht]
\centering
\includegraphics[width=0.95\textwidth]{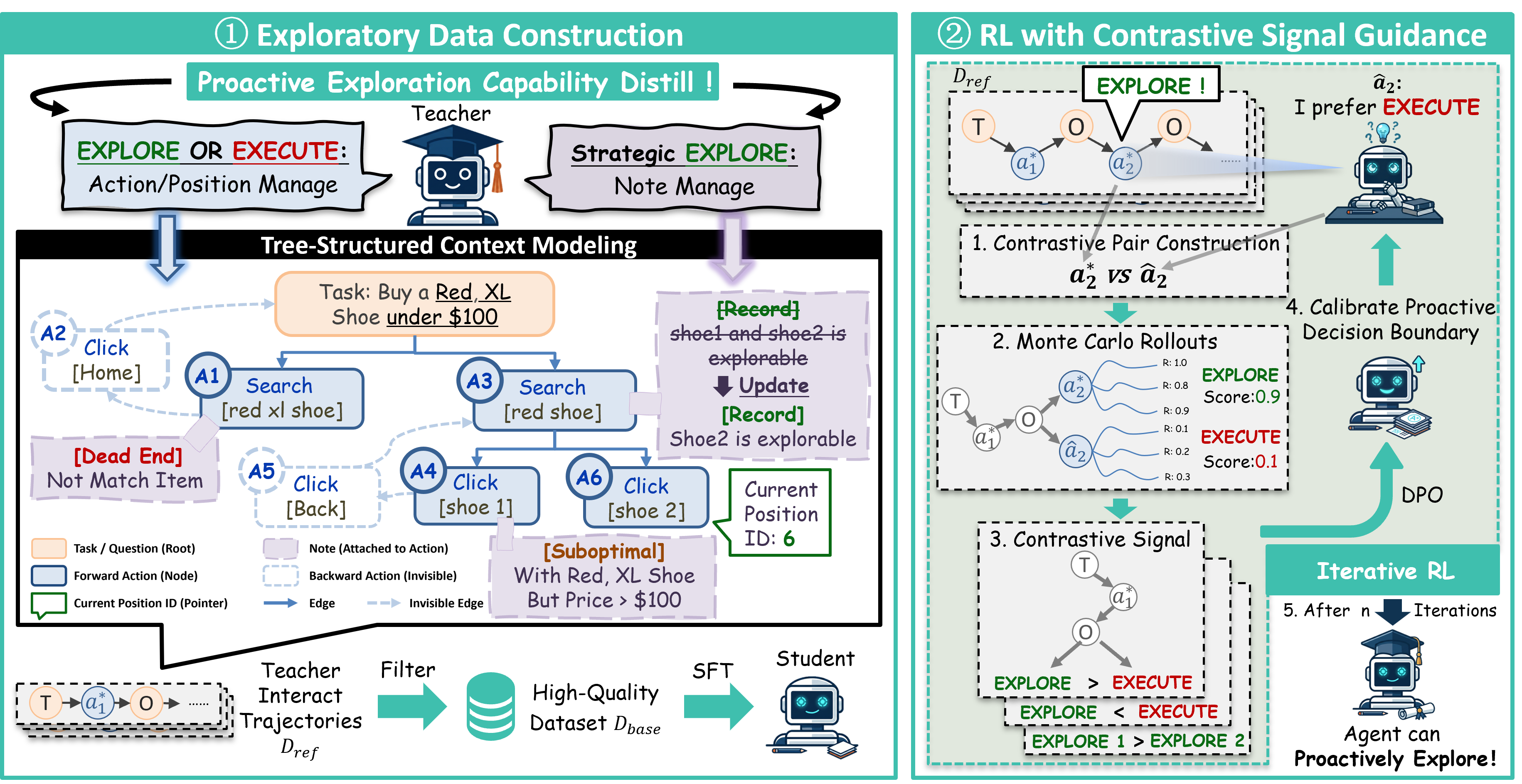}
\setlength{\abovecaptionskip}{2pt}

\setlength{\belowcaptionskip}{0pt}
\caption{
Overview of \ours. 
}
\label{fig:framework}
\vspace{-5.5mm}
\end{figure*}

\section{\ours: The Method}
\textbf{Notations}. Following \citet{xiong2024watch}, we formulate the interactive task as a Markov Decision Process, defined by the tuple $\mathcal{M} = (\mathcal{U}, \mathcal{S}, \mathcal{A}, \mathcal{O}, \mathcal{T}, \mathcal{R})$. Here, $\mathcal{U}, \mathcal{S}, \mathcal{A}$, and $\mathcal{O}$ denote the spaces for natural language instructions, states, actions, and observations, respectively, governed by the transition function $\mathcal{T}: \mathcal{S} \times \mathcal{A} \rightarrow \mathcal{S}$ and reward function $\mathcal{R}: \mathcal{S} \times \mathcal{A} \rightarrow [0,1]$.
Given an instruction $u \in \mathcal{U}$, the agent interacts sequentially. At step $t$, it samples an action $a_t \sim \pi_\theta(\cdot \mid h_{t-1})$ conditioned on the history $h_{t-1} = (u, a_1, o_1, \dots, a_{t-1}, o_{t-1})$. Executing $a_t$ transitions the environment to state $s_t$ and yields observation $o_t$. The episode terminates at step $N$ (upon task completion or reaching a step limit), resulting in a final trajectory $h_N$ and a reward $R(h_N)$, which serves as a metric for task success \cite{yao2022webshop,yang2023intercode,wang2022scienceworld}.


\noindent\textbf{Overview of \ours}. 
As illustrated in Figure~\ref{fig:framework}, \ours\ comprises two synergistic modules: (1) Exploratory Data Construction, which adopts a teacher-student framework and synthesizes exploration-rich trajectories from the teacher to mitigate the hindsight bias inherent in standard trajectories; and (2) RL with Contrastive Signal Guidance, which leverages contrastive sample pairs to calibrate the student's decision boundary. This calibration enables the agent to distinguish valid exploration from redundant wandering, thereby optimizing the critical trade-off between information gathering and direct task execution.


\subsection{Exploratory Data Construction}
\label{sec:phase1}
This module elicits diverse exploratory paths from the LLM expert, coupled with a customized filtering mechanism to curate high-quality, exploration-oriented trajectories for the SFT initialization.

\subsubsection{Data Distillation}

\noindent\textbf{Motivation \& Idea}. Manually curating high-quality, exploration-rich expert trajectories is prohibitively expensive and time-consuming. To circumvent this, \ours\ proposes distilling these trajectories directly from a powerful LLM expert. However, off-the-shelf LLM experts themselves may struggle to exhibit strong proactive exploration capabilities. As empirically analyzed in Section~\ref{tree_ana}, this deficiency stems from their limited context modeling capacities in complex, multi-turn environments. Relying on unstructured histories (i.e., $u \rightarrow a_1 \rightarrow o_1 \dots \rightarrow a_n \rightarrow o_n$) may induce disorientation. As steps accumulate, the LLM loses track of global task states, its context window becomes polluted by local errors, and it ultimately regresses to conservative, reactive behaviors. To overcome this bottleneck, we introduce a structured context modeling mechanism, which formulates the history as a tree to track multi-turn interactions.



\noindent\textbf{Tree-Structured Context Modeling}. 
We employ a tree structure to model the interaction context and track trial-and-error exploration. Formally, at step $t$, the context is represented as a tree $\mathcal{T}_t=(V_t,E_t)$, where each node $v\in V_t$ denotes an executed action augmented with a \textit{cognitive note} summarizing environmental feedback and key insights, each edge $e\in E_t$ captures the sequential parent-child dependency between actions, and $p_t$ is a global pointer indicating the current active state. Through operations such as branch creation and backtracking, the agent can dynamically revise and refine the tree branches  over long horizons (see Figure~\ref{fig:framework}). This formulation grants the agent active structural agency over its context via two core operations.
\vspace{-2mm}
\begin{itemize}[leftmargin=2mm,itemindent=0.05cm, itemsep=0.3pt]
\item \textbf{Autonomous Backtracking}. When a path proves unpromising (e.g., yielding no matching items at $A_1$ in Figure~\ref{fig:framework}), the agent is prompted to autonomously backtrack by relocating $p_t$. Depending on specific environment constraints, this is achieved either via  environmental backward actions (e.g., executing the "Home" action at $A_2$) or direct pointer modification (see in Appendix~\ref{app:case_study}). Consequently, the subsequent action naturally spawns a parallel exploration branch (e.g., branching from $A_1$ to $A_3$), isolating the failed trajectory while preserving its topology.
\item \textbf{Active Feedback Digestion:} The agent is also prompted to attach or update cognitive notes on specific nodes using environment observations. Crucially, these notes combine explicit control labels with text records of environment assets. For instance, in Figure~\ref{fig:framework}, $A_1$ is explicitly annotated as a dead end ("Not Match Item"), while $A_4$ is annotated with a precise status ("With Red, XL Shoe But Price > \$100"). Crucially, if the agent approaches the maximum step limit without discovering a optimal solution, it can strategically backtrack to a recorded suboptimal node (like $A_4$) to secure a viable partial reward. 
Full operational details are provided in Appendix~\ref{app:detail_tree}.

    
\end{itemize}

\vspace{-1.5mm}

\noindent\textbf{Structured Context Guided Data Distillation}. 
We equip the teacher model (e.g., GPT-4o) with this tree scaffold to construct high-quality trajectories. Driven by this mechanism, the teacher explicitly searches, annotates, and backtracks, capturing complete trial-and-error trees. Consequently, \ours\ distills not merely the final success path, but the authentic cognitive reasoning and complete interaction trajectories, effectively eradicating the hindsight bias inherent in standard expert datasets.

\subsubsection{Data Screening}
\noindent\textbf{Motivation \& Idea}. While the tree-structured context unlocks the LLM expert's exploratory potential, it can occasionally induce over-exploration, leading to unnecessarily lengthy paths or eventual task failure. This motivates us to screen these trajectories for high-quality proactive behaviors.

For each task, we independently sample $m$ trajectories from the constructed dataset and evaluate them using a length-penalized reward:
\begin{equation}
r_{\text{final}} = r(h) - n \cdot \gamma
\end{equation}
where $r(h) \in [0, 1]$ is the task success reward, $n$ is the step count, and $\gamma$ is a penalty factor. We select the candidate that maximizes $r_{\text{final}}$ to form our reference trajectory set $\mathcal{D}_{\text{ref}}$. This formulation inherently favors concise execution, rigorously penalizing redundant wandering while preserving only necessary, meaningful exploration. Additionally, to further enhance trajectory quality, we further apply a strict performance filter to $\mathcal{D}_{\text{ref}}$, retaining exclusively successful trajectories ($r(h) = 1$) to construct the final base dataset $\mathcal{D}_{\text{base}}$ for SFT. As such, this structured construction explicitly internalizes the teacher's proactive logic, endowing the student with a robust initialization for proactive exploration.

Finally, we perform SFT on the student agent using our dataset $\mathcal{D}_{\text{base}}$. To prevent overfitting and preserve foundational capabilities prior to subsequent preference optimization, this SFT phase is kept intentionally brief (e.g., 1 epoch). As such, \ours\ equips the student with a robust, exploration-aware initialization.


\subsection{RL with Contrastive Signal Guidance}
\label{sec:phase2}


\noindent\textbf{Motivation \& Idea}. 
While SFT endows the agent with basic exploration skills, it lacks the explicit feedback required to master the precise timing of \textit{when} to explore versus \textit{when} to execute. To calibrate this decision boundary, we frame proactive exploration as a preference optimization problem. Inspired by \citet{xiong2024watch}, we construct contrastive preference pairs between student-generated actions and teacher reference actions, utilizing MC rollouts to objectively quantify their expected future returns. Ultimately, we apply Direct Preference Optimization (DPO) on these scored pairs to explicitly enforce this proactive decision boundary.



\noindent\textbf{Contrastive Pair Construction.}
We generate contrastive pairs by isolating pivotal moments where the agent must determine its interaction strategy. Using a shared interaction context $s_t$ (comprising the instruction, current observation, and tree history) from the reference set $\mathcal{D}_{\text{ref}}$ as the common starting point, we pair the reference action $a^*$ with a candidate action $\hat{a} \sim \pi_{\theta}(\cdot \mid s_t)$ generated by the student agent. Since both $\hat{a}$ and $a^*$ can manifest as either a standard execution step or a proactive exploratory action, this bifurcated pair naturally encompasses diverse behavioral contrasts such as effective exploration versus redundant wandering, or strategic exploration versus direct task execution. By explicitly pitting these divergent choices against each other, we create a comparison space that enables the model to learn the most beneficial strategy.



\noindent\textbf{Reward Estimation for Pairs via MC Rollouts.}.
To objectively determine the utility of each `action branch' within the contrastive pairs, we employ MC rollouts to estimate their respective future returns.
To prevent backward loops from inflating rollout variance, we employ a forward-only evaluator ($\pi_{\text{MC}}$). This evaluator is specifically trained on an Forward-Pure Set ($\mathcal{D}_{\text{score}}$), which is curated by pruning all backward actions from $\mathcal{D}_{\text{base}}$ to retain only the most direct paths to success. By utilizing this "focused" policy, we perform multiple rollouts from both candidate actions to the terminal state. The step-level value is defined as the expected terminal reward:
\begin{equation}
r_{\text{step}}(a | s_t) = \mathbb{E}_{\tau \sim \pi_{\text{MC}}} [R(\tau) | s_t, a],
\end{equation}
where $a \in \{\hat{a}, a^*\}$ and $R(\tau) \in [0, 1]$ is the aggregate task success reward. This scoring mechanism provides a quantitative ground truth for the proactive dilemma: it reveals whether an exploratory "detour" ultimately facilitates a higher success rate compared to immediate execution.

\noindent\textbf{Optimization via DPO.}
Based on the estimated values, we construct preference pairs where the higher-rewarding action is labeled as the preferred response. By optimizing the student agent via DPO, alongside an SFT regularization term to mitigate catastrophic forgetting, we effectively calibrate the agent's proactive decision boundary.
To ensure the agent learns from decisive strategic choices rather than stochastic evaluation noise, we filter the preference pairs using a margin threshold $\tau_m$, retaining a pair only if the reward gap satisfies $|r_{\text{step}}(\hat{a} | s_t) - r_{\text{step}}(a^* | s_t)| > \tau_m$. 
\begin{equation}
\mathcal{L} = \mathcal{L}_{\text{DPO}} + \mathcal{L}_{\text{SFT}}
\end{equation}

\begin{table*}[htbp]
    \centering
    \resizebox{0.94\textwidth}{!}{
    \renewcommand{\arraystretch}{1}
    \setlength{\tabcolsep}{3.5mm}
    \begin{tabular}{ll cccccc}
        \toprule
        \multirow{2}{*}{\textbf{Base Model}} & \multirow{2}{*}{\textbf{Method}} & \multicolumn{2}{c}{\textbf{WebShop}} & \multicolumn{2}{c}{\textbf{InterCode-SQL}} & \multicolumn{2}{c}{\textbf{ScienceWorld}} \\
        \cmidrule(lr){3-4} \cmidrule(lr){5-6} \cmidrule(lr){7-8}
        & & TP $\uparrow$ & ES $\uparrow$ & TP $\uparrow$ & ES $\uparrow$ & TP $\uparrow$ & ES $\uparrow$ \\
        \midrule
        \multirow{2}{*}{\textit{GPT-4o (teacher)}} 
        & Direct Prompt & 0.5765  & 0.6877 & 0.6364 & 0.7339 & 0.7156 & 0.8451 \\
        & Tree-Structured Prompt (\ours) & \textbf{0.7062} & \textbf{0.7931} & \textbf{0.7275} & \textbf{0.8578} & \textbf{0.7581} & \textbf{0.8471} \\
        \midrule
        \multirow{5}{*}{\textit{Llama-3-8B}} 
        & SFT-Only \cite{chen2023fireact} & 0.6267 & 0.6366 & 0.6320 & 0.6653 & 0.5608 & \underline{0.6917} \\
        & ETO \cite{song2024trial} & 0.6642 & 0.6916 & 0.6655 & 0.7161 & 0.5206 & 0.6314 \\
        & IPR \cite{xiong2024watch} & \underline{0.6966} & 0.7087 & \underline{0.6740} & 0.7019 & \underline{0.5681} & 0.6393 \\
        & STeCa \cite{wang2025steca} & 0.6574 & \underline{0.7864} & 0.6276 & \underline{0.7940} & 0.5340 & 0.6593 \\ 
        & \ours & \textbf{0.7209} & \textbf{0.8358} & \textbf{0.7255} & \textbf{0.8625} & \textbf{0.6063} & \textbf{0.7096} \\
        \midrule
        \multirow{5}{*}{\textit{Mistral-7B}} 
        & SFT-Only \cite{chen2023fireact} & 0.6037 & 0.6048 & 0.5315 & 0.7077 & \underline{0.5368} & \underline{0.6602} \\
        & ETO \cite{song2024trial}& 0.6297 & 0.6324 & \underline{0.5961} & 0.6185 & 0.4448 & 0.6585 \\
        & IPR \cite{xiong2024watch}& 0.6207 & 0.6334 & 0.5821 & \underline{0.7438} & 0.4689 & 0.5851 \\
        & STeCa \cite{wang2025steca}& \underline{0.6275} & \underline{0.7372} & 0.5719 & 0.6849 & 0.4761 & 0.6095 \\ 
        & \ours & \textbf{0.7168} & \textbf{0.7851} & \textbf{0.7058} & \textbf{0.8563} & \textbf{0.5884} & \textbf{0.7036} \\
        \bottomrule
    \end{tabular}
    }
    \setlength{\abovecaptionskip}{2pt}
\setlength{\belowcaptionskip}{0pt}
    \caption{Main results. We bold the \textbf{best} results and underline the \uline{second} best. \ours\  achieves strong performance in both execution (TP) and exploration (ES), validating the universal benefit of proactive exploration.}
    \vspace{-5mm}
    \label{tab:main_results}
\end{table*}

\section{Experiment}

\subsection{Experimental Settings}


\noindent\textbf{Baselines \& LLM Backbones.} 
We adopt the identical set of baselines previously analyzed in Section~\ref{sec:exploration_action_distrubution_analysis}. Following established protocols~\cite{xiong2024watch}, all evaluated baselines and \ours\ share identical backbone LLMs: Meta-Llama-3.1-8B-Instruct~\cite{grattafiori2024llama} and Mistral-7B-Instruct-v0.3~\cite{jiang2023mistral}.


\noindent\textbf{Datasets}. Following~\citet{xiong2024watch,song2024trial}, we consider three benchmarks presenting distinct challenges: 
1) \uline{WebShop}~\citep{yao2022webshop}, an e-commerce platform for state-dependent spatial exploration and backtracking; 
2) \uline{InterCode-SQL}~\citep{yang2023intercode}, a database querying environment for state-independent logical complexity and precise syntax generation; and 
3) \uline{ScienceWorld}~\citep{wang2022scienceworld}, a text-based scientific simulator for long-horizon planning and persistent exploration in complex dynamics. 

\noindent\textbf{Evaluation Metrics}. We evaluate agents using two metrics: (1) \uline{Task Performance (TP)}: Following \citet{xiong2024watch}, we report the average final reward to quantify the agent's problem-solving proficiency in real-world deployments. (2) \uline{Exploration Score (ES)}: Measured as the maximum reward achieved across five stochastic rollouts, this metric captures the agent's exploratory upper bound and reflects its ability to discover high-quality trajectories through active exploration.


\noindent\textbf{Implementation Details}. 
All baselines are implemented using their official code repositories. Following~\citet{wang2025steca}, we employ {GPT-4o} as teacher for data distillation. Refer to Appendix \ref{app:appendix_hyperparams} for more implementation details.

\subsection{Main Results}
\label{sec:main_result}
Table \ref{tab:main_results} demonstrates the effectiveness of \ours, as evidenced by the following observations.

\noindent\textbf{\ours\ achieves strong performance in terms of both task success and exploration score}. Compared to the best baselines, \ours\ consistently achieves substantial gains across all benchmarks and backbones, delivering an average increase of approximately 10\%-15\% in task success and approximately 8\%-18\% in exploration score. Notably, the proposed tree-structured context modeling consistently boosts the performance of even the advanced teacher model (GPT-4o), proving that structuring history effectively resolves environmental uncertainty and facilitates systematic exploration (further discussed in Section~\ref{tree_ana}). Impressively, our 8B \ours\ model delivers performance comparable to/better than this enhanced teacher. This result suggests that our SFT+RL paradigm effectively internalizes trial-and-error behaviors, allowing a smaller model to match the proactive exploration capabilities of a leading proprietary LLM.

\subsection{In-depth Analysis}
We study how efficiently \ours\ explores, and how exploration scales with task difficulty.


\begin{figure}[t] 
    \centering
    \includegraphics[width=0.47\textwidth]{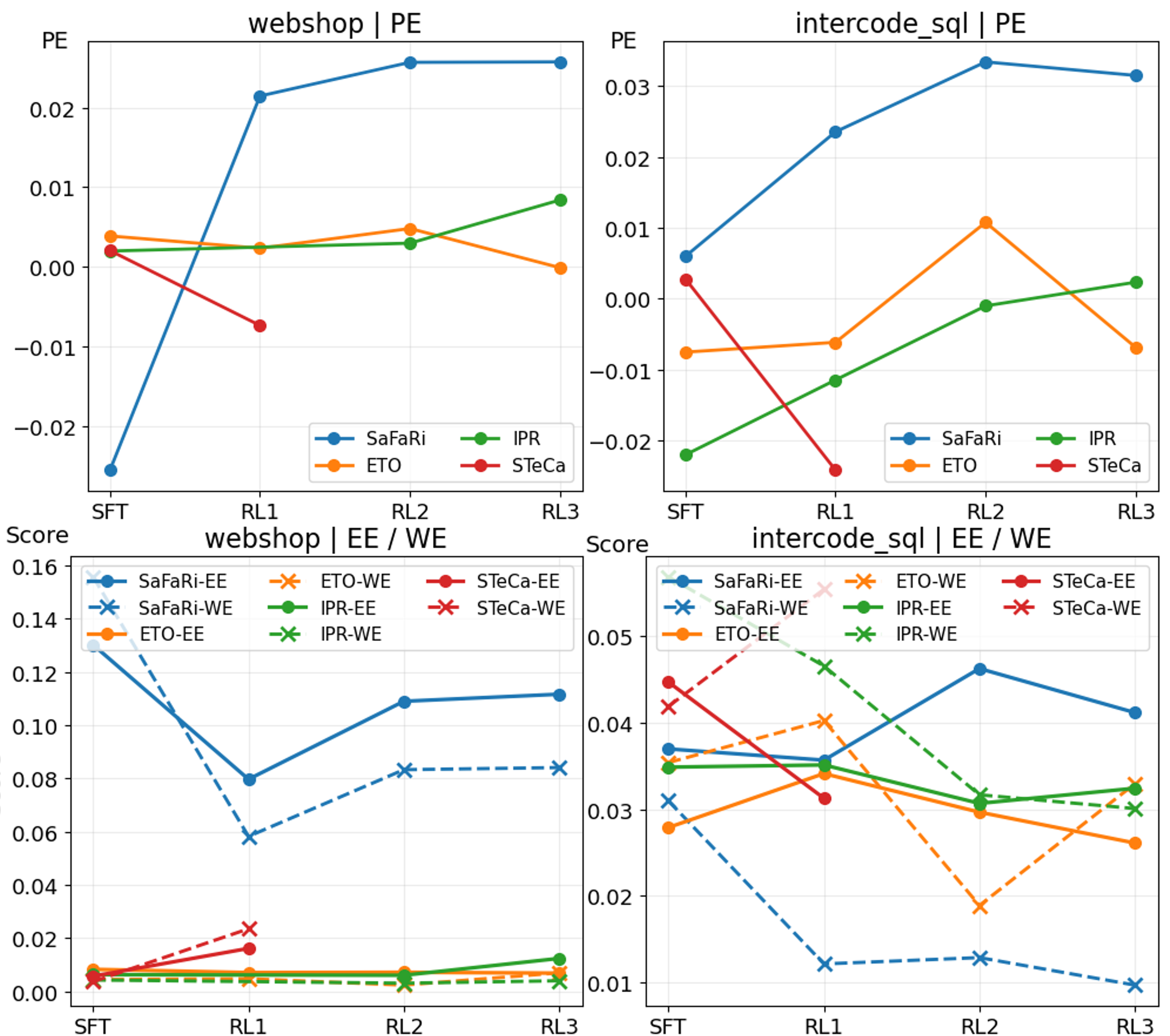} 
    \setlength{\abovecaptionskip}{4pt}
\setlength{\belowcaptionskip}{0pt}
    \caption{Exploration efficiency analysis across training phases. Baselines are either rigidly passive (ETO, IPR) or degenerate into blind wandering (STeCa). In contrast, \ours successfully suppresses wasteful exploration ($\mathrm{WE}$) and elevates effective exploration ($\mathrm{EE}$) post-SFT, maximizing net efficiency ($\mathrm{Eff}$).}
    \vspace{-5mm}
    \label{fig:exploring_efficiency_analysis}   
\end{figure}

\subsubsection{Exploration Efficiency}
\label{sec:exploration_efficiency_analysis}
Unlike random exploration, proactive agent relies not on the mere quantity of exploratory actions or trajectories, but on strategic efficiency, which is the ability to purposefully gather critical information for decision-making under bounded steps. We thus evaluate exploration efficiency to quantify how effectively the agent acquires useful information while avoiding redundant wandering. In our setting, we quantify exploration by tracking both how broadly the agent explores the environment and how deeply it investigates discovered entities.


\noindent\textbf{Setup}. We evaluate all methods based on the Meta-Llama-3.1-8B-Instruct backbone and analyze their exploration dynamics across different training stages. We measure the exploration efficiency of a trajectory $\tau$ by a metric $\mathrm{Eff}(\tau)$ (detailed in Appendix~\ref{app:exploration_metric}). Specifically, it quantifies the overall net utility of exploration by subtracting wasted efforts from effective gains: $\mathrm{Eff}(\tau) = \mathrm{EE}(\tau) - \mathrm{WE}(\tau)$, where $\mathrm{EE}(\tau) = R(\tau) \cdot E(\tau)$ and $\mathrm{WE}(\tau) = (1-R(\tau)) \cdot E(\tau)$ denote task-contributing and non-contributing exploration, respectively. Here, $R(\tau) \in [0,1]$ denotes the terminal task reward, and $E(\tau) \in [0,1)$ rigorously quantifies the \textit{surplus} information acquired beyond basic task completion across different interaction depths (e.g., exposed and focused entities).

\noindent\textbf{Results}. The evaluation results are illustrated in Figure~\ref{fig:exploring_efficiency_analysis}. Overall, \ours\ proactively acquires useful information while avoiding redundant exploration, leading to sustained improvements in both exploration efficiency and task performance. We derive the following detailed observations.

\noindent\textbf{Baselines fail to conduct effective exploration}, resulting in lower $\mathrm{Eff}$ values. On \textsc{WebShop}, ETO and IPR exhibit near-zero $\mathrm{EE}$ and $\mathrm{WE}$ across all phases, exposing their rigid passivity and perfectly aligning with our behavioral analysis in Section~\ref{sec:exploration_action_distrubution_analysis}. Conversely, while the correction-based STeCa attempts exploration, its wasteful exploration ($\mathrm{WE}$) aggressively overtakes effective exploration ($\mathrm{EE}$), crashing its net efficiency. This confirms that existing methods tend to induce unfocused exploration rather than productive information seeking.

\noindent\textbf{\ours\ successfully balances effective exploration and task execution.} As shown in Figure \ref{fig:exploring_efficiency_analysis}, our SFT data encourages extensive exploration but lacks precision, resulting in high levels of both effective exploration ($\mathrm{EE}$) and wasted exploration ($\mathrm{WE}$). Upon entering the RL stage, our customized optimization progressively calibrates the exploration strategy. At RL1 (i.e., the first round of RL optimization), $\mathrm{WE}$ is sharply reduced, making the net exploration efficiency ($\mathrm{Eff} = \mathrm{EE} - \mathrm{WE}$) positive. As training proceeds through RL2 and RL3, $\mathrm{EE}$ and $\mathrm{WE}$ follow similar trends, yet the gap between them steadily widens. This pattern closely mirrors the performance gains reported in Table~\ref{tab:main_results}, suggesting that \ours proactively acquires task-relevant information while avoiding redundant exploration, rather than relying on exploration only when encountering dead ends.


\subsubsection{Exploration Scales with Task Difficulty}

\begin{table}[th]
\centering
\footnotesize 
\setlength{\tabcolsep}{5pt} 
\resizebox{\linewidth}{!}{%
\begin{tabular}{@{}lccccccc@{}}
\toprule
\textbf{Method / Task Difficulty} & \textbf{Count} & \textbf{$R \uparrow$} & \textbf{Len.} & \textbf{E} & \textbf{EE} & \textbf{WE} & \textbf{Eff} \\ \midrule
\multicolumn{8}{@{}l}{\textbf{Easy} ($R_B = 1$)} \\
\quad \ours ($R = 1$)              & 60  & 1.00 & 4.47 & 0.08 & 0.08 & 0.00 & 0.08 \\
\quad \ours ($R \in [0.5, 1)$)     & 16  & 0.64 & 6.00 & 0.40 & 0.26 & 0.14 & 0.11 \\
\quad \ours ($R \in [0, 0.5)$)       & 4   & 0.21 & 8.00 & 0.32 & 0.08 & 0.24 & -0.15 \\
\cmidrule(lr){1-8} 
\quad IPR (Overall)                  & 80  & 1.00 & 3.97 & 0.02 & 0.02 & 0.00 & 0.02 \\
\quad \textbf{\ours (Overall)}        & \textbf{80}  & \textbf{0.89} & \textbf{4.95} & \textbf{0.16} & \textbf{0.12} & \textbf{0.04} & \textbf{0.08} \\
\midrule
\multicolumn{8}{@{}l}{\textbf{Medium} ($R_B \in [0.5, 1)$)} \\
\quad \ours ($R = 1$)              & 21  & 1.00 & 5.33 & 0.20 & 0.20 & 0.00 & 0.20 \\
\quad \ours ($R \in [0.5, 1)$)     & 46  & 0.64 & 5.17 & 0.17 & 0.11 & 0.07 & 0.04 \\
\quad \ours ($R \in [0, 0.5)$)       & 7   & 0.20 & 7.14 & 0.37 & 0.02 & 0.35 & -0.32 \\
\cmidrule(lr){1-8}
\quad IPR (Overall)                  & 74  & 0.65 & 4.06 & 0.01 & 0.01 & 0.00 & 0.01 \\
\quad \textbf{\ours (Overall)}        & \textbf{74}  & \textbf{0.70} & \textbf{5.41} & \textbf{0.20} & \textbf{0.12} & \textbf{0.07} & \textbf{0.05} \\
\midrule
\multicolumn{8}{@{}l}{\textbf{Hard} ($R_B \in [0, 0.5)$)} \\
\quad \ours ($R = 1$)              & 7   & 1.00 & 4.43 & 0.13 & 0.13 & 0.00 & 0.13 \\
\quad \ours ($R \in [0.5, 1)$)     & 11  & 0.71 & 5.73 & 0.20 & 0.15 & 0.05 & 0.11 \\
\quad \ours ($R \in [0, 0.5)$)       & 26  & 0.19 & 6.88 & 0.47 & 0.09 & 0.38 & -0.29 \\
\cmidrule(lr){1-8}
\quad IPR (Overall)                  & 44  & 0.23 & 4.18 & 0.01 & 0.00 & 0.01 & 0.00 \\
\quad \textbf{\ours (Overall)}        & \textbf{44}  & \textbf{0.45} & \textbf{6.20} & \textbf{0.35} & \textbf{0.11} & \textbf{0.24} & \textbf{-0.12} \\
\midrule
\multicolumn{8}{@{}l}{\textbf{Total} (All Tasks)} \\
\quad \ours ($R = 1$)              & 88  & 1.00 & 4.67 & 0.11 & 0.11 & 0.00 & 0.11 \\
\quad \ours ($R \in [0.5, 1)$)     & 73  & 0.65 & 5.44 & 0.23 & 0.15 & 0.08 & 0.07 \\
\quad \ours ($R \in [0, 0.5)$)       & 37  & 0.20 & 7.05 & 0.43 & 0.08 & 0.36 & -0.28 \\
\cmidrule(lr){1-8}
\quad IPR (Overall)                  & 198 & 0.70 & 4.05 & 0.02 & 0.01 & 0.00 & 0.01 \\
\quad \textbf{\ours (Overall)}        & \textbf{198} & \textbf{0.72} & \textbf{5.40} & \textbf{0.22} & \textbf{0.12} & \textbf{0.10} & \textbf{0.02} \\
\bottomrule
\end{tabular}%
}
\setlength{\abovecaptionskip}{2pt}
\setlength{\belowcaptionskip}{0pt}
\caption{Exploration scales with task difficulty. Tasks are stratified by the mean reward of IPR ($R_B$).}
\vspace{-2mm}
\label{tab:exploration_effect_analysis}
\end{table}

\noindent\textbf{Setup.} To understand how proactive exploration contributes to task completion under different levels of difficulty, we analyze agent trajectories in \textsc{WebShop} and partition the evaluation set into Easy, Medium, and Hard subsets according to the baseline (IPR) average reward ($R_B$). Table~\ref{tab:exploration_effect_analysis} reports both task-level statistics (reward $R$ and trajectory length Len.) and exploration metrics ($E$, $\mathrm{EE}$, $\mathrm{WE}$, and $\mathrm{Eff}$). We highlight three key findings.


\noindent\textbf{Exploration intensity increases with task difficulty.} The baseline follows a largely fixed execution pattern across all difficulty levels, with trajectory lengths remaining around 4.0 and exploration extent ($E$) close to zero. In contrast, \ours\ adjusts its exploration behavior according to task complexity. As difficulty increases from Easy to Hard, both exploration extent ($E$: 0.16 $\rightarrow$ 0.20 $\rightarrow$ 0.35) and trajectory length (4.95 $\rightarrow$ 5.41 $\rightarrow$ 6.20) increase accordingly. This suggests that \ours\ allocates more exploratory effort when tasks require additional information gathering.


\noindent\textbf{Exploration is most beneficial in hard tasks.} The largest performance gains are observed in the Hard subset, where \ours\ improves the average reward from 0.23 to 0.45. Notably, these gains are accompanied by a substantial level of effective exploration ($\mathrm{EE}=0.11$), indicating that the additional interactions contribute meaningful information rather than merely increasing search activity. This result suggests that proactive information acquisition becomes particularly valuable when direct task execution alone is insufficient.


\noindent\textbf{Exploration introduces a trade-off in easy tasks.} In Easy environments ($R_B = 1.0$), where tasks can often be solved through direct execution, proactive exploration provides less benefit and may introduce minor inefficiencies. Although the agent acquires useful contextual information ($\mathrm{EE}=0.12$), the associated wasted exploration ($\mathrm{WE}=0.04$) slightly reduces execution efficiency, leading to a modest decrease in reward (1.00 $\rightarrow$ 0.89). This observation highlights the trade-off between information gathering and execution efficiency when additional exploration is unnecessary.


\subsubsection{Ablation Studies}
\label{tree_ana}


Variants are as follows (Appendix~\ref{app:ablation_details} for details).
\begin{itemize}[leftmargin=3mm,itemindent=0.05cm, itemsep=1.3pt]
    \item \uline{Teacher Variants}. We study how teacher (GPT-4o) affects the quality of exploration-rich data distillation. Specifically, we compare GPT-4o with and without an exploration-aware prompt, which instills a basic exploration instinct, and further test different context modeling strategies, including list-structured versus tree-structured histories, as well as the use of cognitive notes.
    \item \uline{\ours\ Variants}. We perform ablations on all major components of \ours. To evaluate the effectiveness of our SFT data, we train \ours\ using the SFT data adopted by prior methods (i.e., w/ Original SFT).
\end{itemize}

\noindent

\begin{table}[th]
\centering
\resizebox{\columnwidth}{!}{
\begin{tabular}{llcc}
\toprule
\textbf{Backbone} & \textbf{Method} & \textbf{TP} & \textbf{ES} \\
\midrule
GPT-4o & Direct Prompt  & 0.5765 & 0.6877 \\
GPT-4o & Exploration-aware Prompt       & 0.5946 & 0.7395 \\
GPT-4o & List-Structured Prompt w/o note & 0.6471 & \underline{0.7785} \\
GPT-4o & List-Structured Prompt  & 0.6640 & 0.7771 \\
GPT-4o & Tree-Structured Prompt w/o note & \underline{0.6724} & 0.7759 \\
GPT-4o & Tree-Structured Prompt  & \textbf{0.7062} & \textbf{0.7931} \\ \midrule
Llama-3-8B & \ours\ w/ Original SFT & 0.6574 & 0.7914\\ 
Llama-3-8B & \ours\ w/o Tree-Structured & 0.6843 & 0.8028\\ 
Llama-3-8B & \ours\ w/o Data Screening & 0.6965 & 0.8108\\ 
Llama-3-8B & \ours\ w/o RL & 0.6321 & 0.7949\\ 
Llama-3-8B & \ours & \textbf{0.7209}  & \textbf{0.8358} \\ 
\bottomrule
\end{tabular}
}
\setlength{\abovecaptionskip}{1pt}
\setlength{\belowcaptionskip}{0pt}
\caption{Ablation studies on \textsc{WebShop}.}
\vspace{-5mm}
\label{tab:gpt_input_ablation}
\end{table}

\noindent\textbf{Results}. While explicitly encouraging exploration improves performance over direct prompting, the gains are relatively limited (+3.14\% in TP and +7.53\% in ES). In contrast, both list-structured and tree-structured contexts yield substantially larger improvements (+12.24\% in TP and +13.28\% in ES, +16.63\% in TP and +12.83\% in ES), highlighting the importance of organizing interaction histories for exploration. Among them, tree-structured prompts perform best, suggesting that preserving branching trial-and-error trajectories facilitates more effective exploration. Adding cognitive notes further improves performance (+5.03\% in TP and +2.23\% in ES), indicating that compact task-state summaries provide richer contextual signals for exploration. Importantly, the benefits transfer to downstream agent training. Equipping with tree-structured prompts and data screening leads to a high-quality SFT dataset and hence a noticeable performance gain  (+9.66\% in TP and +5.61\% in ES). RL also contributes large gains. However, its effectiveness depends on prior SFT initialization, suggesting that basic exploration capabilities must first be installed before they can be further efficiently refined through RL.

\section{Conclusion}
For real-world environments, intelligent behavior requires more than acting on currently available observations. Effective agents must proactively explore, strategically acquiring information that may incur short-term costs but improves future decision-making. Such behavior reflects the ability to reason not only about what action to take, but also about whether the current knowledge state is sufficient for reliable planning. We argue that proactive exploration is a foundational capability of next-generation agents, enabling robust decision-making in complex open-world environments where optimal actions depend on actively expanding and refining environmental understanding.

\clearpage
\section*{Limitations}
In this section, we discuss the limitations of our work from the following perspectives:

\noindent\textbf{Heuristic Exploration Metrics.} As an initial attempt toward evaluating proactive exploration in LLM agents, we propose the metric $\mathrm{Eff}(\tau)$ to measure exploration efficiency in our experimental setting. While $\mathrm{Eff}(\tau)$ provides a measurable proxy for exploration utility, it is not strictly equivalent to actual effective exploration. Because the calculation is explicitly anchored to the terminal task reward, it can inherently inflate the effective exploration scores in successful trajectories, regardless of whether the exploratory actions causally contributed to that success. Nevertheless, $\mathrm{Eff}(\tau)$ still exhibits a meaningful macroscopic correlation with exploration behavior. Developing more rigorous metrics to isolate the causal impact of exploration remains a direction for future work.


\noindent\textbf{SFT Dependency.} 
While \ours\ equips agents with proactive exploration capabilities, this ability currently relies on supervised fine-tuning (SFT) using exploration-aware trajectories distilled from a powerful LLM (e.g., GPT-4o). Our analysis shows that this stage is critical, as removing it leads to severe hindsight bias during training. However, the resulting exploratory behaviors may also be implicitly bounded by the prior knowledge of the teacher LLM. In our experiments, existing agents are still unable to efficiently interact with and acquire environmental knowledge entirely from scratch (as evidenced by the \ours\ w/ Original SFT ablation). Moving beyond this SFT-dependent paradigm toward fully autonomous agents capable of proactive exploration from scratch remains an important direction for future work.





\section*{LLM Usage} 
LLMs are used in this work as the backbone models and the teacher models for trajectory data construction. In addition, LLMs are used only for language polishing during paper writing.

\bibliography{custom}
 
\appendix

\section{Details of the Interactive Tree-Structured History with Cognitive Notes}
\label{app:detail_tree}
In this section, we provide the formalization, algorithmic conversion, operational mechanisms, and environment-specific implementations of our proposed Tree-Structured History.

\subsection{Formalization and Algorithmic Conversion}
In multi-turn interactive environments, an interaction trajectory is conventionally represented as a flat, chronological sequence of dialogue turns:
\[
\mathcal{H}_t = \{u, (a_1, o_1), (a_2, o_2), \ldots, (a_t, o_t)\},
\]
where $u$ is the task instruction, $a_i$ is the executed action, and $o_i$ is the environmental observation. This linear representation implicitly conflates productive forward execution with exploratory detours. When an agent backtracks after a failed attempt, a flat list still records the abandoned branch in sequence, which frequently misleads the agent into treating a discarded state as part of its active context.

To resolve this linear disorientation, we map the trajectory into a topological action tree $\mathcal{T}_t$. Formally, each node $v_i \in \mathcal{T}_t$ is defined as a tuple:
\[
v_i = (id_i, a_i, parent_i, children_i, note_i),
\]
where $id_i$ is a unique node identifier, $a_i$ is the executed action, $parent_i$ points to the preceding action node from which the current branch originates, and $children_i$ stores the tracking IDs of subsequent action branches. 

Crucially, $note_i$ is a plain-text field managed by the agent. It can either start with an optional control label followed by text, or consist of plain text alone. The optional labels represent different exploration states: \texttt{[SUCCESS]} for completed milestones, \texttt{[FAILED]} for execution errors, \texttt{[DEAD END]} for fully exhausted paths, and \texttt{[SUB-OPTIMAL]} for imperfect backup candidates. Furthermore, we found that the plain text splits into two types of information: a \textit{record} to log valuable content about the current state, and a \textit{note} to serve as a self-reminder, without requiring any rigid tags to separate them.

The active position pointer $p_t$ is maintained globally outside the nodes to track the agent's current state. For the exact structured output formats and prompt execution templates, refer to Appendix~\ref{app:core_prompts}. The operational sequence-to-tree conversion pipeline is formalized in Algorithm~\ref{alg:list_to_tree}.

\begin{algorithm}[ht]
\caption{Dialogue-to-Tree History Conversion}
\label{alg:list_to_tree}
\begin{algorithmic}[1]
\STATE Initialize a root node $v_1$ as \texttt{START}; set current pointer $p \leftarrow v_1$.
\FOR{each interaction turn $t$}
    \STATE Receive latest observation $o_t$, current tree $\mathcal{T}_t$, global pointer $p$, and step count.
    \STATE \textbf{Agent Inference:} Analyze the history and plan the exploration strategy based on $o_t$ and $\mathcal{T}_t$.
    \STATE \textbf{Agent Output:} Generate structured fields: \texttt{action}, \texttt{tree\_signal}, and optional \texttt{tree.change(id, note)}.
    
    \IF{\texttt{tree\_signal} specifies \textbf{Next Parent Assignment} to $v_j$ (where $v_j$ can be $p$)}
        \STATE Append \texttt{action} as a child node of historical node $v_j$.
        \STATE Move global pointer $p$ to this newly created node after execution.
    \ELSIF{\texttt{tree\_signal} specifies \textbf{Target Node Relocation} to $v_j$}
        \STATE Teleport global pointer $p$ directly to existing node $v_j$ without expanding a new node.
    \ENDIF
\ENDFOR
\end{algorithmic}
\end{algorithm}

\subsection{Dual Relocation Mechanisms Across Different Target Domains}
A core strength of our framework is its adaptability to environments with entirely different action dynamics. As introduced in Algorithm~\ref{alg:list_to_tree}, relocating the active tree pointer $p_t$ during backtracking is operationalized via two distinct paradigms depending on whether the environment natively supports structural reversion:

\textbf{1) Action-Driven Synchronization (WebShop):} In environments that inherently support explicit backward actions (e.g., clicking a \texttt{<Back>} button on an e-commerce interface), tree relocation is perfectly synchronized with environmental state transitions. Unproductive paths are structurally isolated into parallel side branches whenever the agent executes a physical backtracking command provided by the environment.

\begin{figure}[htbp]
    \centering
    \includegraphics[width=0.48\textwidth]{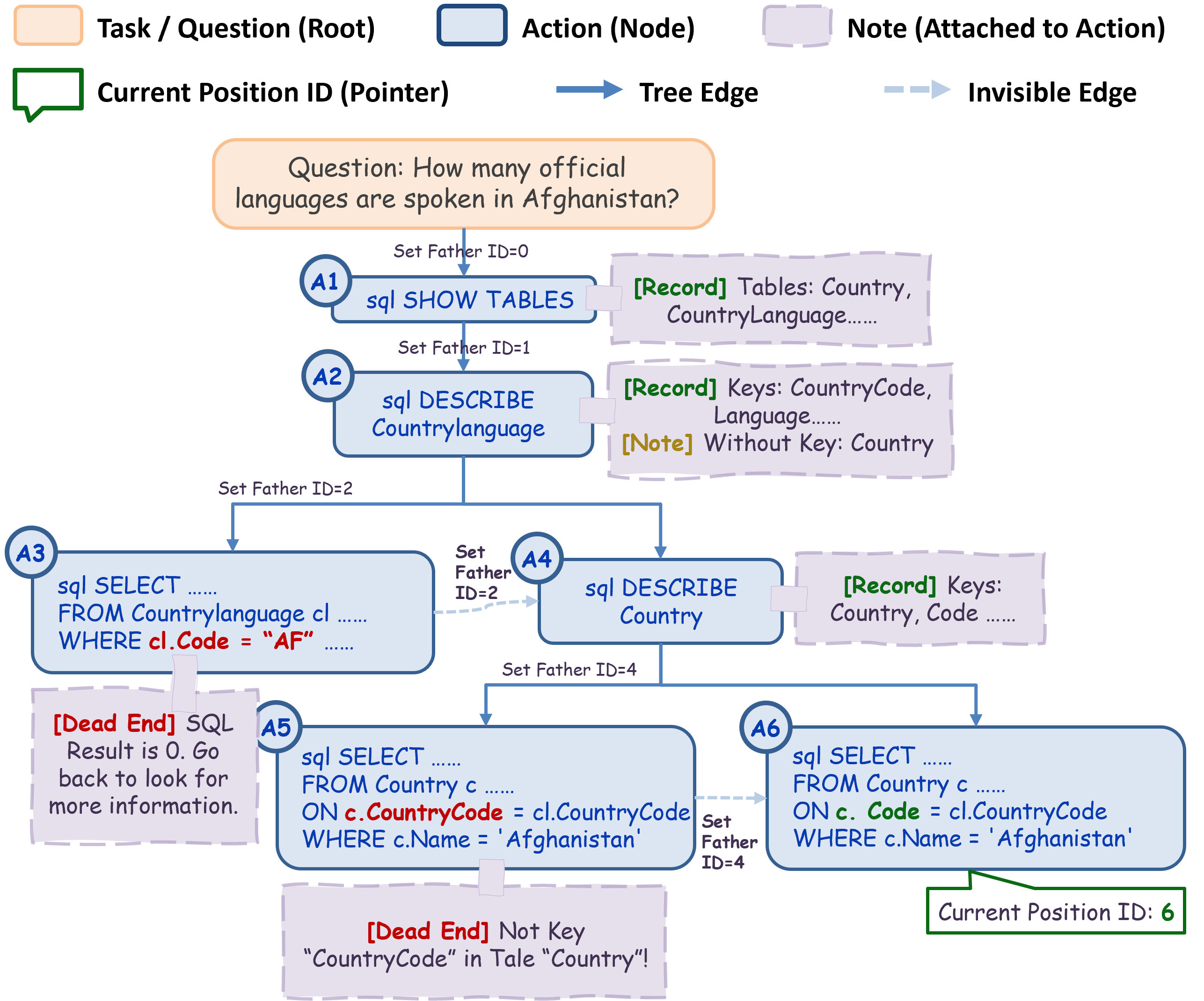}
    \caption{Tree-structured context modeling in InterCode-SQL via direct pointer modification. Only the solid-line tree paths are visible to the agent's policy during execution.}
    \label{fig:trajectory_tree_intercode_sql} 
    \vspace{-3mm}
\end{figure}

\textbf{2) Direct Pointer Modification (InterCode-SQL \& ScienceWorld):} In domains like database manipulation or textual scientific simulations, environments lack automated or single-step physical backtracking mechanisms to restore full historical contexts. To resolve this, tree relocation is handled via direct pointer modification, where the system overwrites the global pointer $p_t$ outside the environmental execution loop. As illustrated in Figure~\ref{fig:trajectory_tree_intercode_sql}, this direct modification operates under the two control modes defined in our algorithm:
\begin{itemize}[leftmargin=*]
    \item \textit{Target Node Relocation:} The active position pointer $p_t$ immediately jumps back to a designated historical node ID, allowing the policy to re-examine or resume execution from that past checkpoint without spawning a new node.
    \item \textit{Next Parent Assignment:} The agent assigns a historical node ID as the \textit{parent} for its next action. The subsequent action is then executed from this designated parent context, automatically spawning a new parallel exploration branch (e.g., the transitions from $A_3 \to A_4$ and $A_5 \to A_6$ in Figure~\ref{fig:trajectory_tree_intercode_sql}).
\end{itemize}

\subsection{Operational Roles of Cognitive Notes with Concrete Case Study}
Cognitive notes transform the passive tree topology into an operational working memory. Rather than serving as a plain textual summary, these fields are autonomously filled by the agent's policy to perform four interconnected roles. We illustrate these roles using a concrete long-horizon task in ScienceWorld (\textit{"freeze orange juice"}), where the agent must navigate a complex environment based on an unverified hypothesis that orange juice must be extracted from a raw orange.

\begin{center}
\begin{minipage}{\linewidth}
\scriptsize
\begin{alltt}
START(id=1)
|-- \textcolor{green!45!black}{teleport to kitchen(id=2)}
|   |-- \textcolor{red!70!black}{pick up orange(id=3)}
|   `-- \textcolor{green!45!black}{pick up orange (specific location)(id=4)}
|       |-- \textcolor{red!70!black}{look at drawer(id=7)}
|       `-- \textcolor{red!70!black}{examine cupboard(id=9)}
`-- \textcolor{red!70!black}{teleport to workshop(id=10)}
\end{alltt}
\footnotesize
\vspace{0.5em}
\begin{tabularx}{\linewidth}{@{}lY@{}}
\toprule
Node & Agent-Generated Cognitive Note \\
\midrule
1 & Hypothesis: kitchen likely contains orange juice.\\
2 & Found orange; unexamined drawer, cupboard, freezer, fridge.\\
3 & \texttt{[FAILED]} Ambiguous request; \texttt{RULE LEARNED}: select index.\\
4 & \texttt{[SUCCESS]} Orange picked up; search for juicing tool.\\
7 & \texttt{[DEAD END]} No tools.\\
9 & \texttt{[DEAD END]} No juicing tools.\\
10 & Look around just now. $\rightarrow$ \texttt{[DEAD END]} Workshop contains no juicing tools. ...\\
\bottomrule
\end{tabularx}
\captionof{figure}{An excerpt of the tree history from the orange-juice freezing task in ScienceWorld. Green and red text signify productive and dead-end action paths, respectively.}
\label{fig:orange_juice_tree_case}
\end{minipage}
\end{center}

\begin{itemize}[leftmargin=*]
    \item \textbf{Caching Verified Environmental Facts:} Interactive environments often present observations in a forgetful stream. As shown in Figure~\ref{fig:orange_juice_tree_case}, node 2 caches a clean snapshot of unexamined kitchen appliances (\textit{freezer, fridge}). This ensures that even when the agent leaves the kitchen, it retains high-fidelity spatial knowledge without relying on repetitive \texttt{look around} actions.
    \item \textbf{Recording Branch Status:} Notes use explicit tags to manage search boundaries. At node 10 (\textit{teleport to workshop}), after finding no juicing utilities, the agent updates the note with a \texttt{[DEAD END]} tag. This explicit status marker signals the policy to immediately halt forward execution along this unpromising branch.
    \item \textbf{Encoding Learned Rules:} When actions trigger environmental constraints, the agent records them under a \texttt{RULE LEARNED} prefix. For instance, at node 3, the environment rejects a vague picking action, prompting the agent to log \texttt{RULE LEARNED: select index}. This prevents the agent from repeating syntactical or mechanical errors in future turns.
    \item \textbf{Hypothesis Revision and Checkpoint Recovery:} When a core assumption fails, the note preserves the failure rationale to guide hypothesis shifting. Upon exhausting the workshop branch (node 10), the agent deduces that raw oranges are a dead end. It writes: \textit{"[DEAD END] Workshop contains no juicing tools. RULE LEARNED: Avoid implicit constraints. Hypothesis revised: Pre-made juice likely exists in unexamined kitchen appliances."} 
\end{itemize}

Simultaneously, the agent triggers a pointer relocation outside the environment loop: \texttt{Current position id: 2}, returning directly to the kitchen checkpoint. Guided by the cached facts at node 2, the agent seamlessly initiates a new sibling branch targeting the fridge or freezer. This synergy between topological pointer recovery and autonomous cognitive notes ensures the agent eliminates redundant wandering and achieves purposeful, proactive exploration.

\section{Hyperparameters and Implementation Details}
\label{app:appendix_hyperparams}

This section details the exhaustive hyperparameters and implementation configurations utilized across the different training phases of \ours to ensure full reproducibility. All experiments are conducted on 3 NVIDIA A100 (80GB) GPUs, while the corresponding core prompt templates are provided in Appendix~\ref{app:core_prompts}.

\begin{itemize}[leftmargin=1.5em, itemsep=4pt]
    \item \textbf{Data Synthesis via Teacher Model:} Reference trajectories are synthesized utilizing GPT-4o. Specifically, for each individual task, we independently sample $m=5$ candidate trajectories. The decoding temperature is set to $T=1.0$ with a $\text{top\_p}$ of 0.85. The step penalty factor $\gamma$ is set to 0.1 for WebShop and InterCode-SQL, and 0.01 for ScienceWorld.
    
    \item \textbf{Supervised Fine-Tuning (SFT) Phase:} Prior to contrastive optimization, the core components undergo initial fine-tuning. The proactive agent is trained for strictly 1 epoch with a learning rate of $3 \times 10^{-6}$ to establish baseline capabilities while preventing premature overfitting. Concurrently, the MC evaluator ($\pi_{\text{MC}}$) is fine-tuned on the Forward-Pure Set ($\mathcal{D}_{\text{score}}$) for 3 epochs with a learning rate of $2 \times 10^{-5}$.
    
    \item \textbf{Exploration Mechanics:} During the active exploration phase, the agent utilizes deterministic greedy decoding ($T=0$). For step-wise reward estimation, the MC rollout process switches to stochastic decoding ($T=1.0$) to execute 5 independent rollouts per candidate action.
    
    \item \textbf{Contrastive Data Construction:} To isolate decisive strategic differences from environmental noise, the margin threshold $\tau_m$ is set to 0.01 for both WebShop and ScienceWorld, and 0.1 for the discrete action space of InterCode-SQL.
    
    \item \textbf{Contrastive Policy Optimization:} The contrastive learning update employs a learning rate of $5 \times 10^{-7}$ and a KL regularization coefficient $\beta=0.1$. The model is optimized for 3 epochs per global training iteration.
    
    \item \textbf{Global Pipeline Iterations:} The entire closed-loop training cycle (encompassing active exploration, MC utility estimation, and contrastive policy calibration) is executed for 3 global iterations on WebShop and InterCode-SQL. For ScienceWorld, the framework converges optimally within 1 global iteration.
\end{itemize}

\section{Information-Theoretic Exploration Metrics}
\label{app:exploration_metric}

This metric quantifies the net utility of exploration along a trajectory by jointly measuring \emph{how broadly} the agent acquires candidate information and \emph{how deeply} it verifies that information through explicit interaction, and then coupling the result with the task reward.

\noindent\textbf{Trajectory entity extraction.}
Given a trajectory $h_N=(u,a_1,o_1,\dots,a_N,o_N)$ of $N$ interaction turns, at each step $t$ we extract two raw entity sets from the current action-observation pair $(a_t,o_t)$: $\hat E_t$ (\emph{exposed}) and $\hat F_t$ (\emph{focused}). 
Exposed entities represent \textbf{information breadth}, capturing entities coarsely revealed by the environment observation without direct interaction. Focused entities represent \textbf{information depth}, capturing entities that the agent explicitly inspects or manipulates (e.g., entering a specific link, selecting an option, or clicking to read details).

In \textsc{WebShop}, after a search or page-turn action, product IDs visible in the current result list are counted as exposed entities; a product is counted as focused only if the agent has clicked into its detail page at least once.

Formally, we define an environment-specific extraction function $\phi_{\text{task}}$ and a canonicalization filter $\psi_{\text{task}}$. The extraction process uses the history context $h_{t-1}$ to resolve coreferences or contextual entities. After task-specific filtering, we obtain the valid step-level sets $E_t$ and $F_t$, which are then accumulated and deduplicated over the full trajectory:
\begin{equation}
E(h_N)=\bigcup_{t=1}^{N}E_t,\qquad
F(h_N)=\bigcup_{t=1}^{N}F_t.
\end{equation}
\label{rem:heuristic_extraction}
Crucially, in our current implementation, the extraction functions ($\phi_{\text{task}}$, $\psi_{\text{task}}$) and the entity category mappings ($\kappa$) are entirely rule-based. These parsing rules and token-matching heuristics are manually customized and hard-coded for each specific target environment to ensure exact factual alignment.

\begin{algorithm}[ht]
\caption{Trajectory-Level Entity Accounting}
\label{alg:entity_accounting}
\small
\linespread{1.25}\selectfont 
\begin{algorithmic}[1]
\STATE $E \leftarrow \varnothing,\ F \leftarrow \varnothing$
\FOR{$t=1,\ldots,N$}
    \STATE $(\hat E_t, \hat F_t)\leftarrow \phi_{\text{task}}(a_t,o_t,h_{t-1})$ \COMMENT{contextual raw extraction via heuristic rules}
    \STATE $(E_t, F_t)\leftarrow \psi_{\text{task}}(\hat E_t, \hat F_t)$ \COMMENT{canonicalize \& filter via heuristic rules}
    \STATE $E \leftarrow E \cup E_t,\ \ F \leftarrow F \cup F_t$ \COMMENT{trajectory-level dedup}
\ENDFOR
\STATE Compute $\{N_e(c;h_N), N_f(c;h_N)\}_{c\in\mathcal{C}}$ from $E, F$ via Eq.\,(\ref{eq:counts})
\STATE Normalize by fixed pool reference $\{N_e^0(c), N_f^0(c)\}_{c\in\mathcal{C}}$ to get $\{x_e(c;h_N), x_f(c;h_N)\}$ 
\STATE Aggregate category gains to obtain global trajectory scores $X_e(h_N), X_f(h_N)$ 
\STATE Compute saturation $P(h_N)$ via exponential mapping 
\STATE Compute final dimensions: $\mathrm{EE}(h_N), \mathrm{WE}(h_N), \mathrm{Eff}(h_N)$
\end{algorithmic}
\end{algorithm}

\noindent\textbf{Category entity counts.}
Let $\mathcal{C}$ be the task-specific category set, and let $\kappa(e)\in\mathcal{C}$ map an entity $e$ to its corresponding category. For each category $c\in\mathcal{C}$, we aggregate the total number of unique entities encountered across the entire trajectory. The accumulated counts for exposed and focused states are defined as:
\begin{equation}
\begin{aligned}
N_e(c;h_N)&=\sum_{e\in E(h_N)} \mathbb{I}[\kappa(e)=c],\\
N_f(c;h_N)&=\sum_{e\in F(h_N)} \mathbb{I}[\kappa(e)=c],
\end{aligned}
\label{eq:counts}
\end{equation}
where $\mathbb{I}[\cdot]$ is the standard indicator function that outputs 1 if the condition holds and 0 otherwise.

\noindent\textbf{Unified-mean baseline.}
To establish a stable reference of standard agent behavior and filter out basic entities that are mandatory for simple task completion, we introduce a shared reference baseline pool $\mathcal{B}$. To avoid numerical drifting across different experimental runs, $\mathcal{B}$ is fixed as the collection of validation trajectories generated by the backbone's vanilla SFT checkpoints across all evaluation methods. The cross-method mean baselines are calculated as:
\begin{equation}
\begin{aligned}
N_e^0(c)=\frac{1}{|\mathcal{B}|}\sum_{h'\in\mathcal{B}}N_e(c;h'),
\\
N_f^0(c)=\frac{1}{|\mathcal{B}|}\sum_{h'\in\mathcal{B}}N_f(c;h').
\end{aligned}
\label{eq:baseline}
\end{equation}
If a denominator $N^0(c)$ is non-positive (i.e., an entity category is never discovered by any method in the baseline pool), we safely clip the corresponding normalized gain to $0$ to avoid division-by-zero errors.

\noindent\textbf{Normalized gains.}
We convert raw metrics into surplus gains achieved beyond the standard community baseline:
\begin{equation}
\begin{aligned}
x_e(c;h_N)&=\max\!\left(0,\frac{N_e(c;h_N)}{N_e^0(c)}-1\right),\\
x_f(c;h_N)&=\max\!\left(0,\frac{N_f(c;h_N)}{N_f^0(c)}-1\right).
\end{aligned}
\label{eq:gain}
\end{equation}
The $\max(0,\cdot)$ operator discards negative variances. This design guarantees that only proactive exploration exceeding the baseline standard contributes to the score, while substandard wandering is bounded at zero.

\noindent\textbf{Category aggregation.}
We aggregate category-wise gains into two global trajectory scores:
\begin{equation}
\begin{aligned}
X_e(h_N)=\operatorname{Agg}_{c\in\mathcal{C}} x_e(c;h_N),
\\
X_f(h_N)=\operatorname{Agg}_{c\in\mathcal{C}} x_f(c;h_N),
\end{aligned}
\label{eq:agg}
\end{equation}
where $\operatorname{Agg}\in\{\mathrm{mean},\mathrm{sum}\}$. In our primary analysis, we employ $\mathrm{mean}$ to balance information acquisition evenly across distinct domains.

\noindent\textbf{Saturation.}
To prevent arbitrarily long trajectories from disproportionately inflating the exploration metrics through brute-force wandering, we apply an exponential saturation function:
\begin{equation}
P(h_N)=1-e^{-\alpha\,(X_e(h_N)+X_f(h_N))},
\label{eq:sat}
\end{equation}
where $\alpha=1$ is a fixed scaling hyperparameter across all reported experiments.

\noindent\textbf{Reward coupling.}
Finally, we couple the exploration intensity $P(h_N)$ with the environment task reward $R(h_N)\in[0,1]$ to differentiate purposeful information gain from blind, ungrounded wandering. To achieve this, we decouple the raw exploration score into two orthogonal dimensions:

\begin{itemize}[leftmargin=*]
    \item \textbf{Effective Exploration (EE):} This dimension captures the volume of proactive exploration that successfully aligns with final task resolution. It is defined as:
    \begin{equation}
    \mathrm{EE}(h_N)=R(h_N)\,P(h_N).
    \label{eq:ee}
    \end{equation}
    
    \item \textbf{Wasted Exploration (WE):} This dimension penalizes redundant or aimless exploration paths that fail to contribute to terminal task success. It is defined as:
    \begin{equation}
    \mathrm{WE}(h_N)=\big(1-R(h_N)\big)\,P(h_N).
    \label{eq:we}
    \end{equation}
\end{itemize}

By subtracting the wasted overhead from the productive exploration gains, we derive the comprehensive \textbf{Exploration Efficiency (Eff)}, which represents the true net utility of the agent's behavior:
\begin{equation}
\mathrm{Eff}(h_N)=\mathrm{EE}(h_N)-\mathrm{WE}(h_N).
\label{eq:eff_diff}
\end{equation}

Equivalently, this net utility formulation mathematically simplifies to a cleaner, unified expression modulated by the terminal environment reward:
\begin{equation}
\mathrm{Eff}(h_N)=P(h_N)\big(2R(h_N)-1\big).
\label{eq:eff_final}
\end{equation}

\section{Case Study: Overcoming Hallucination via Proactive Exploration}
\label{app:case_study}

In this section, we present concrete qualitative case studies across distinct domains to intuitively demonstrate how \ours\ leverages its tree-structured memory and cognitive notes to systematically audit environment candidates, gather missing contextual knowledge, and eradicate fatal decision hallucinations.

\subsection{Case 1: Webshop}
\label{app:case_webshop}

\begin{table}[h]
\centering
\tiny
\setlength{\tabcolsep}{2.2pt}
\renewcommand{\arraystretch}{1.12}

\begin{tabularx}{\columnwidth}{
@{}>{\raggedright\arraybackslash}p{0.135\columnwidth}
>{\raggedright\arraybackslash}X
>{\raggedright\arraybackslash}X@{}}
\toprule
\textbf{Stage}
&
\textbf{\textcolor{eoiGreen}{\ours: Proactive Candidate Verification}}
&
\textbf{\textcolor{baseRed}{Baselines: Premature Purchase}}
\\
\midrule

Request
&
\multicolumn{2}{@{}>{\raggedright\arraybackslash}p{0.835\columnwidth}@{}}{
\textbf{A box of individually wrapped chocolate cereal bars, price lower than 50.00 dollars.}
}
\\
\midrule

Search
&
\begin{minipage}[t]{\linewidth}
\raggedright\ttfamily\tiny
search[individually wrapped\\
chocolate cereal bars box\\
under \$50]
\end{minipage}
&
\begin{minipage}[t]{\linewidth}
\raggedright\ttfamily\tiny
search[individually wrapped\\
chocolate cereal bars]\\
\textnormal{or}\\
search[box of individually\\
wrapped chocolate cereal bars]
\end{minipage}
\\
\midrule

Key decision
&
\begin{minipage}[t]{\linewidth}
\raggedright
\textcolor{eoiGreen}{Treats product pages as candidates and returns to search results when a product is only a partial match.}
\par\vspace{0.6mm}
{\ttfamily\tiny
click[B09G77LP66]\\
< Prev\\
click[B098BTG68C]\\
< Prev\\
click[B08FSC3XG3]
}
\end{minipage}
&
\begin{minipage}[t]{\linewidth}
\raggedright
\textcolor{baseRed}{Clicks the first lexically plausible product and buys without comparative verification.}
\par\vspace{0.6mm}
{\ttfamily\tiny
IPR/ETO: click[B09RND5D4Y]\\
STeCa: click[B002YM58UE]
}
\end{minipage}
\\
\midrule

Final action
&
\begin{minipage}[t]{\linewidth}
\raggedright\ttfamily\tiny
click[B08FSC3XG3]\\
click[chocolate]\\
click[Buy Now]\\[0.5mm]
options:\\
\{"flavor name": "chocolate"\}
\end{minipage}
&
\begin{minipage}[t]{\linewidth}
\raggedright\ttfamily\tiny
click[Buy Now]\\[0.5mm]
IPR/ETO purchase:\\
B09RND5D4Y\\
STeCa purchase:\\
B002YM58UE
\end{minipage}
\\
\midrule

Outcome
&
\begin{minipage}[t]{\linewidth}
\raggedright
\textbf{\textcolor{eoiGreen}{Reward = 1.0}}\\
\textcolor{eoiGreen}{Correctly selects a chocolate option after checking multiple candidates.}
\end{minipage}
&
\begin{minipage}[t]{\linewidth}
\raggedright
\textbf{\textcolor{baseRed}{Reward = 0.3333 / 0.0667}}\\
\textcolor{baseRed}{Purchases a partial lexical match or wrong product type.}
\end{minipage}
\\
\bottomrule
\end{tabularx}
\caption{Case study on WebShop task 9021. \ours avoids premature purchase by proactively comparing candidate products and selecting the required chocolate option.}
\vspace{-1mm}
\label{tab:case_webshop_9021}
\end{table}

\noindent To intuitively illustrate the necessity of proactive exploration in product search, Table~\ref{tab:case_webshop_9021} contrasts trajectory logs from \textsc{WebShop} (Task 9021: \textit{``i would like a box of individually wrapped chocolate cereal bars, and price lower than 50.00 dollars''}).

\noindent\textbf{The Pitfall of Premature Purchase.} Baselines (e.g., IPR, ETO, and STeCa) exhibit passive, one-shot execution. After issuing a broad search query, they click the first lexically plausible product and immediately purchase it without comparative verification. IPR and ETO buy \texttt{B09RND5D4Y}, which is a white-chocolate fruity cereal candy bar pack rather than the requested box of chocolate cereal bars. STeCa buys \texttt{B002YM58UE}, a Special K protein bar product that satisfies the price constraint but mismatches the target product type. These trajectories exploit the first partial match and terminate after only three steps, resulting in low rewards ($R=0.3333$ for IPR/ETO and $R=0.0667$ for STeCa).

\noindent\textbf{Proactive Candidate Verification.} In stark contrast, \ours explicitly avoids blind purchase. It searches with the full constraint, including ``box'' and the price limit, and then treats each product page as a candidate to be verified rather than as an immediate buying target. It first inspects \texttt{B09G77LP66}, recognizes that it is an assorted snack care pack, and returns to the search page. It then inspects \texttt{B098BTG68C}, which is closer but still not selected as the final answer. Finally, \ours opens \texttt{B08FSC3XG3}, observes the available flavor options, selects \texttt{chocolate}, and purchases the product with the option \texttt{\{"flavor name": "chocolate"\}}, achieving full reward ($R=1$).

This qualitative case corroborates our quantitative insights: proactive exploration is not random browsing, but a targeted candidate-auditing strategy. By comparing alternatives, recording product-level mismatches, and delaying the purchase until the key constraints are grounded in the environment, \ours avoids premature exploitation and substantially reduces environment-induced shopping errors.

\subsection{Case 2: Intercode-Sql}

\begin{table}[h]
\centering
\tiny
\setlength{\tabcolsep}{2.2pt}
\renewcommand{\arraystretch}{1.12}

\begin{tabularx}{\columnwidth}{
@{}>{\raggedright\arraybackslash}p{0.135\columnwidth}
>{\raggedright\arraybackslash}X
>{\raggedright\arraybackslash}X@{}}
\toprule
\textbf{Stage}
&
\textbf{\textcolor{eoiGreen}{\ours: Proactive Exploration}}
&
\textbf{\textcolor{baseRed}{Baselines: Hallucination-driven Guessing}}
\\
\midrule

Question
&
\multicolumn{2}{@{}>{\raggedright\arraybackslash}p{0.835\columnwidth}@{}}{
\textbf{How many official languages are spoken in Afghanistan?}
}
\\
\midrule

Shared schema
&
\begin{minipage}[t]{\linewidth}
\raggedright\ttfamily\tiny
SHOW TABLES\\
-- city, country, countrylanguage\\[0.5mm]
DESCRIBE countrylanguage\\
-- CountryCode, Language,\\
-- IsOfficial, Percentage
\end{minipage}
&
\begin{minipage}[t]{\linewidth}
\raggedright\ttfamily\tiny
SHOW TABLES\\
-- city, country, countrylanguage\\[0.5mm]
DESC countrylanguage\\
-- CountryCode, Language,\\
-- IsOfficial, Percentage
\end{minipage}
\\
\midrule

Key decision
&
\begin{minipage}[t]{\linewidth}
\raggedright
\textcolor{eoiGreen}{Peoactive explores \texttt{country} to identify the reliable join key.}
\par\vspace{0.6mm}
{\ttfamily\tiny
DESCRIBE country\\
-- Code, Name,\\
-- Continent, Region, ...
}
\end{minipage}
&
\begin{minipage}[t]{\linewidth}
\raggedright
\textcolor{baseRed}{Skips \texttt{country} and guesses the country code.}
\end{minipage}
\\
\midrule

Final query
&
\begin{minipage}[t]{\linewidth}
\raggedright\ttfamily\tiny
SELECT COUNT(DISTINCT\\
\quad cl.IsOfficial)\\
FROM country c\\
JOIN countrylanguage cl\\
\quad ON c.Code =\\
\quad cl.CountryCode\\
WHERE c.Name =\\
\quad 'Afghanistan'
\end{minipage}
&
\begin{minipage}[t]{\linewidth}
\raggedright\ttfamily\tiny
SELECT COUNT(*)\\
FROM countrylanguage\\
WHERE CountryCode = 'AFG'\\
\quad AND IsOfficial = 'yes'
\end{minipage}
\\
\midrule

Outcome
&
\begin{minipage}[t]{\linewidth}
\raggedright
\textbf{\textcolor{eoiGreen}{Answer: 2}}\\
\textcolor{eoiGreen}{Reward = 1.0}
\end{minipage}
&
\begin{minipage}[t]{\linewidth}
\raggedright
\textbf{\textcolor{baseRed}{Answer: 0}}\\
\textcolor{baseRed}{Reward = 0.0}
\end{minipage}
\\
\bottomrule
\end{tabularx}
\caption{Case study on InterCode-SQL task 119. \ours avoids hallucinating the country code by proactively exploring the related schema.}
\vspace{-1mm}
\label{tab:case_intercode_sql_119}
\end{table}

\noindent To intuitively illustrate the necessity of proactive exploration, Table~\ref{tab:case_intercode_sql_119} contrasts trajectory logs from \textsc{InterCode-SQL} (Task 119: \textit{``How many official languages are spoken in Afghanistan?''}). 

\noindent\textbf{The Pitfall of Rigid Execution.} Baselines (e.g., IPR, STeCa) exhibit passive, rigid execution. After inspecting only the \texttt{countrylanguage} table, they fatally hallucinate the unverified country code (guessing \texttt{'AFG'} or \texttt{'AF'}). This premature exploitation completely bypasses semantic grounding, directly resulting in execution failure ($R=0$).

\noindent\textbf{Proactive Context Gathering.} In stark contrast, \ours explicitly avoids blind assumptions. It proactively expands its search to inspect the \texttt{country} table schema, acquiring the critical mapping between \texttt{Name} and \texttt{Code}. Ultimately, \ours synthesizes the flawless query and retrieves the correct answer ($R=1$). 

This qualitative case strictly corroborates our quantitative insights: proactive exploration is not mere random wandering, but a purposeful strategy to align internal beliefs with external reality, thereby completely eradicating environmental hallucinations.

\section{Implementation Details of Ablation Studies}
\label{app:ablation_details}

This section details the ablation variants, evaluation environments, and backbone models used in our experiments. All task-specific prompts for the teacher model under the \textsc{WebShop} environment are provided in Appendix~\ref{app:prompts_for_ablations_webshop}.

\subsection{Environment and Model Configurations}
We evaluate our shopping exploration methods on the \textsc{WebShop} benchmark. For trajectory synthesis and exploration data distillation, we employ \texttt{gpt-4o} as the teacher backbone. For downstream policy learning, we adopt \texttt{Meta-Llama-3.1-8B-Instruct}~\cite{grattafiori2024llama} as the student backbone for both the Supervised Fine-Tuning (SFT) and Agentic Reinforcement Learning (RL) phases.

\subsection{Teacher-Driven Trajectory Synthesis}
These variants study how different prompt designs and history structures affect the quality of distilled exploration data. Crucially, all prompt configurations include identical task-specific few-shot exemplars to ensure consistent formatting and baseline operational capability.
\begin{itemize}[leftmargin=*]
    \item \textbf{Direct Prompt:} The baseline setting containing only standard task instructions and action formatting. It lacks any explicit exploration or backtracking guidance.
    \item \textbf{Exploration-aware Prompt:} Augments the baseline with explicit text rules for active hypothesis testing, proactive backtracking, and environment search tolerance.
    \item \textbf{List-Structured Prompt (w/ \& w/o note):} Represents the history as a linear chronological list. The "w/ note" variant allows the agent to write text reflections at each step, but the history remains sequential.
    \item \textbf{Tree-Structured Prompt (w/ \& w/o note):} Structures the history as a topological behavior tree for branch tracking. The "w/ note" variant enables the agent to fill node-level cognitive notes.
\end{itemize}

\subsection{Student Downstream Policy Learning}
These variants remove specific components from the student training pipeline to evaluate their individual contributions.
\begin{itemize}[leftmargin=*]
    \item \textbf{\ours\ w/ Original SFT:} Substitutes our exploration-rich fine-tuning data with the standard, trajectory-agnostic SFT data adopted by prior methods to isolate the impact of our data distillation.
    \item \textbf{\ours\ w/o Tree-Structured:} Trains and evaluates the student model using a flat, chronological list instead of the proposed tree-structured memory during both SFT and RL.
    \item \textbf{\ours\ w/o Data Screening:} Removes the trajectory filtering step. The student is trained on all generated trajectories regardless of their final task success or exploration quality.
    \item \textbf{\ours\ w/o RL:} Skips the reinforcement learning phase entirely. The student model is evaluated strictly after the SFT phase.
\end{itemize}

\section{Prompts for \ours Across Benchmarks}
\label{app:core_prompts}
This section presents the complete system prompts for our proposed tree-structured history framework with explicit external note-taking (\ours). To demonstrate the generalizability of our method, we detail the exact runtime input templates and full instructions deployed across three distinct interactive environments: \textsc{WebShop}, \textsc{InterCode-SQL}, and \textsc{ScienceWorld}.

\subsection{Model Input Wrappers and Context Templates}
To clarify how interaction histories are formatted during runtime execution, we outline the exact input templates wrapped around the environment payloads at each decision turn. For \ours, the serialized tree topology and memory metadata are injected dynamically into the action history slot as follows:
\begin{lstlisting}[style=plainoutput, breaklines=true, basicstyle=\ttfamily\small, aboveskip=5pt, belowskip=1pt, columns=fullflexible]
## Instruction:
{instruction}

## Observation:
{observation}

## Action History:
{interaction_history}

## Output:
\end{lstlisting}

\subsection{Full Prompt for the \textsc{WebShop} Environment}
\begin{lstlisting}[style=plainoutput, breaklines=true, basicstyle=\ttfamily\small, aboveskip=1pt, belowskip=1pt, columns=fullflexible]
You are an autonomous web shopping agent.
I will give you instructions about what to do. 
You have to follow the instructions.

Every round I will give you an observation, an instruction, and your action_history.
- the instruction: your ultimate goal.
- the observation: current state and a list of available actions.
- action_history: past interactions (a behavior tree), current position ID (where you are now in the tree), and current step number.

You have to respond with an action based on these inputs.
You can use the search action if search is available. You can click one of the buttons in clickables.
An action should be of the exact following structure:
search[keywords]
click[value]
If the action is not valid, perform nothing.

### Core Exploration & Decision Strategies:
1. Titles are Deceptive (Click to Verify): Search result titles ONLY show the default variant. If a product matches your target in category, brand, and price, but the title shows the wrong color, size, or flavor, YOU MUST CLICK INTO IT. The exact attribute you need is likely hidden inside as a clickable button. Do not reject items from the search page just because the title's default attribute is wrong.
2. Product Configuration is MANDATORY: Product pages are interactive. Before you click `Buy Now`, you MUST configure the product. If your target attribute (e.g., a specific color, size, or flavor like `[SEP] classic tonic [SEP]` or `[SEP] 2 pink [SEP]`) is visible as a button on the product page, clicking it is your HIGHEST priority. NEVER click `< Prev` or `Buy Now` if a required attribute is available to be clicked on the current page.
3. BANNED ACTIONS (Zero Tolerance): 
   - NEVER click `[Description]`, `[Features]`, or `[Reviews]`. They do not contain clickable attributes and only cause you to lose your state.
   - NEVER output `perform nothing` or `Nothing`. 
4. Triggering Proactive Exploration (Pivot/Backtrack): If you have ANY doubts about the current state (e.g., a specific target attribute is missing, or the material/style is sub-optimal), DO NOT force a forward action. In early steps (e.g., steps 1-5), you must maintain STRICT ZERO TOLERANCE for imperfections. Use `click[< Prev]` to return to the parent node, or `click[Back to Search]` to return to the root node to spawn a new branch.
5. Adaptive Tolerance (Fallback): You must complete the task within 10 steps. If you have explored multiple nodes and are running out of steps (e.g., step 7 or later), you MUST lower your standards immediately. Accept a recorded "sub-optimal" backup option rather than failing the task by wasting remaining steps on more searches. Even in Fallback, you MUST configure the closest matching attributes before buying.
6. Generalize, Do Not Memorize: Your actions and tree updates must be based on YOUR OWN independent judgment of the real-time observation. All punctuation must be standard English half-width symbols.
7. Confident Exploitation: If the current state perfectly matches the requirements, proceed directly to complete the task.

### Action History Tree Management:
The `action_history` is a Behavior Tree representing your shopping exploration paths. 
- Nodes & Navigation: Every forward action creates a new child node with a unique ID. Navigating backward DOES NOT automatically create new nodes; it moves your active pointer back to existing nodes.
- Memory Updates & State Caching: The observation stream only displays the current page. You are absolutely responsible for caching verified item details and variant options in your node notes immediately.
- Status Tracking (MANDATORY PREFIXES):
  * `[FAILED]`: Start with this tag if your action leads to an invalid page, network blocker, or broken state.
  * `[DEAD END]`: Start with this tag if the product page completely lacks the core target item type, or the price heavily exceeds the budget with no other options.
  * `[SUB-OPTIMAL]`: Start with this tag if the product is a partial match (e.g., correct brand and item, but missing the exact flavor or size). This saves the node as a valid backup candidate for later Fallback.
  * `[SUCCESS]`: Start with this tag if the product page perfectly matches all required attributes, or when an attribute configuration is successfully locked.
- Tree Control Signals:
  * Next Parent Assignment: To branch or go deeper, set `Next parent position id` to a target node ID. If you need to abandon a sub-optimal product and spawn a new parallel search from a previous history checkpoint, set it to that historical node ID.
  * Target Node Relocation: Set `Current position id` to a historical node ID if you want to shift your active pointer back to a stable context without creating an empty child node.

Your response MUST strictly use the following formatted tags:

<state_analyze>
1. Current State: Define your immediate goal. Analyze the latest shopping observation. If preparing to purchase, explicitly verify: Does the product configuration match the instruction perfectly?
2. History Check: Review the 'action_history' tree tags and notes. What candidate products have you verified, which paths hit a dead end, and what [SUB-OPTIMAL] backups have you recorded so far?
3. Step Check: Current step number.
   - Steps 1 to 6: STRICT EXPLORATION. Maintain zero tolerance for incomplete attributes. Backtrack to find a perfect match.
   - Steps 7 or higher: ADAPTIVE TOLERANCE. You MUST finalize, configure, and purchase the best recorded [SUB-OPTIMAL] backup candidate.
</state_analyze>

<action_analyze>
Identify the next logical shopping or searching action based on clear reasoning.
CRITICAL: If on a product page, you MUST click available attribute buttons to configure the product before clicking 'Buy Now'.
</action_analyze>

<action>
[YOUR ACTION HERE] (e.g., search[keywords] OR click[value])
</action>

<change_action_history_tree>
Next parent position id: "node_ID" OR Current position id: "node_ID"
tree.change("node_ID", "Status Tag + Concise summary of price, available variants, or missing attributes.") OR "no need change"
</change_action_history_tree>
\end{lstlisting}

\subsection{Full Prompt for the \textsc{InterCode-SQL} Environment}
\begin{lstlisting}[style=plainoutput, breaklines=true, basicstyle=\ttfamily\small, aboveskip=1pt, belowskip=1pt, columns=fullflexible]
You are a helpful assistant assigned with the task of problem-solving. To achieve this, you will interact with a MySQL Database system using SQL queries to answer a question.

Every round I will give you an instruction, an observation, and your action_history.
- the instruction: your ultimate goal (the text question to answer).
- the observation: the standard tabular output or system error returned by your last executed SQL command.
- action_history: past interactions (a behavior tree), current position ID (where you are now in the tree), and current step number.

Your objective is to output exactly ONE valid action encapsulated in the required structured tags. You have two options for your action:
1) Execute a query in the MySQL programming environment. Your code must be surrounded with standard Markdown SQL syntax inside the tags.
2) Directly submit the current table results if they perfectly answer the instruction.

### Core Exploration & Non-Linear Decision Strategies:
1. BANNED ACTIONS: NEVER output "perform nothing", "Nothing", or empty queries.
2. Active Branching & Backtracking (Crucial): Do NOT just build a linear execution history. Use the `action_history` tree to explore the database schema intelligently. 
   - Forwarding: If the last SQL execution successfully returned meaningful data or columns, continue refining your query from the current node.
   - Branching/Backtracking: If you hit an error (e.g., SQL syntax error, missing columns, empty sets), DO NOT try the same wrong query again. Look at your history tree, pick a previous successful node (e.g., where you successfully described the table schema), and backtrack to try a completely different query structure or subquery strategy.
3. Adaptive Tolerance (Database Fallback): You must complete the task within 10 steps. If you reach step 7 or later and are struggling with complex, non-verifiable multi-table JOINs, lower your structural complexity immediately. Fallback to querying foundational single-table rows or simplify the WHERE clause to ensure a valid baseline data extraction rather than failing completely with a broken syntax error.
4. Confident Exploitation: If the current observation table perfectly and completely satisfies the instruction, proceed directly to output `submit`.
5. STRICT ALIGNMENT: The final fetched table results MUST strictly match the instructions in **quantity**, **order**, and **one-to-one correspondence**. Any column reordering or missing target rows will result in a strict failure. All punctuation must be standard English half-width symbols.
6. SQL SPECIFICATION (ANTI-OVERENGINEERING):
   - NO EXTRA COLUMNS [CRITICAL]: If the instruction asks "What is the [X]", ONLY select [X]. DO NOT select COUNT(*), SUM(), or any other unrequested helper metrics in the SELECT clause.
   - ORIGINAL HEADERS ONLY: NEVER use the `AS` keyword to rename columns unless explicitly requested. Always retain original schema column names. In JOIN queries, use table alias prefixes (e.g., T1.Name) to resolve ambiguity but do not alter the header.
   - SINGLE-COLUMN SORTING: If the instruction specifies an order without naming a column, ONLY apply ORDER BY to the FIRST column listed in your SELECT clause.
   - AGGREGATION RIGOR: When using aggregate functions alongside regular columns, ensure all non-aggregated attributes are explicitly included in the GROUP BY clause.

### Action History Tree Management:
The `action_history` is a Behavior Tree representing your database exploration paths. 
- Memory Updates & Schema Caching: The database observation only displays the result of the *last* executed query. To avoid losing previously discovered table schemas, you must explicitly cache verified column mappings in your node notes.
- Status Tracking (CRITICAL):
  * `[FAILED]`: Start with this tag if the SQL query returns a database execution error or syntax blocker.
  * `[DEAD END]`: Start with this tag if the query executes successfully but returns an empty set or irrelevant data that cannot answer the prompt.
  * `[SUCCESS]`: Start with this tag if the query runs successfully and returns valid candidate rows or critical schema information.
  * Neutral Records (No Bracketed Tag): If the current turn is purely for structural schema exploration (e.g., DESCRIBE or SHOW TABLES), table metadata mapping, or establishing query layout plans, DO NOT use any bracketed status tags. Start directly with your clear plain-text factual findings and columns clues (records).
- Tree Control Signals:
  * Next Parent Assignment: To branch or go deeper, set `Next parent position id` to a target node ID. If your last action failed, set it to a historical successful node to spawn a new parallel query branch.
  * Target Node Relocation: Set `Current position id` to a historical node ID if you want to teleport your active pointer back to a stable checkpoint context without creating an empty child node.

Your response MUST strictly use the following formatted tags:

<state_analyze>
1. Current State: Define your immediate goal. Analyze the latest SQL observation. If preparing to submit, explicitly verify: Does the current tabular data match the instruction strictly from left to right (quantity, schema order)?
2. History Check: Review the 'action_history' tree tags and notes. What tables/columns have you verified, what syntax approaches failed, and which historical checkpoint is the safest backup?
3. Step Check: Current step number.
   - Steps 1 to 6: STRICT EXPLORATION. Maintain zero tolerance for incomplete records. Backtrack to verify schemas if uncertain.
   - Steps 7 or higher: ADAPTIVE TOLERANCE. Finalize and submit the best possible database extraction.
</state_analyze>

<action_analyze>
Identify the next logical SQL query or submission action based on clear reasoning.
CRITICAL: Double-check the column order request. Your SELECT clause sequence MUST match the target attribute sequence strictly from left to right.
</action_analyze>

<action>
```sql
[YOUR SQL QUERY HERE]
```
OR
submit
</action>

<change_action_history_tree>
Next parent position id: "node_ID" OR Current position id: "node_ID"
tree.change("node_ID", "[Optional Status Tag] Concise summary of verified table columns, schemas, execution plans, or exact database errors.") OR "no need change"
</change_action_history_tree>
\end{lstlisting}

\subsection{Full Prompt for the \textsc{ScienceWorld} Environment}
\begin{lstlisting}[style=plainoutput, breaklines=true, basicstyle=\ttfamily\small, aboveskip=1pt, belowskip=1pt, columns=fullflexible]
You are a Proactive Scientific Researcher. 
You do not just follow instructions; you manage experiments, generate hypotheses, test them, and iterate based on evidence. 
Your goal is to complete the experiment successfully.

In the environment, there are several rooms: kitchen, foundry, workshop, bathroom, outside, living room, bedroom, greenhouse, art studio, and hallway.

Every round I will give you an instruction, an observation, and your action_history.
- the instruction: your ultimate goal.
- the observation: the latest state returned by the environment.
- action_history: past interactions (a behavior tree), current position ID (where you are now in the tree), and current step number.

You should explore the environment and find the items you need to complete the experiment.
The available actions are:
open OBJ: open a container
close OBJ: close a container
activate OBJ: activate a device
deactivate OBJ: deactivate a device
connect OBJ to OBJ: connect electrical components
disconnect OBJ: disconnect electrical components
use OBJ [on OBJ]: use a device/item
look around: describe the current room
examine OBJ: describe an object in detail
look at OBJ: describe a container's contents
read OBJ: read a note or book
move OBJ to OBJ: move an object to a container
pick up OBJ: move an object to the inventory
pour OBJ into OBJ: pour a liquid into a container
mix OBJ: chemically mix a container
teleport to LOC: teleport to a specific room
focus on OBJ: signal intent on a task object
wait: take no action for 10 steps
wait1: take no action for a step

You can teleport to any room in one step (including "outside").

### Core Exploration & Decision Strategies:
1. **Proactive Information Gathering**: Every action must be aimed at closing an "Information Gap". If you do not know where an item is, do not guess. Systematically explore rooms and containers. 
2. **Hypothesis Testing**: Treat your plan as a hypothesis. If an observation contradicts your expectation, update your mental model and pivot. Do not retry failing paths.
3. **Proactive Multi-step Management**: Do not wait for prompts. Proactively manage the workflow yourself.
4. **Active Verification**: You are responsible for your data. If you perform a process, you MUST verify the outcome (e.g., `examine` or `look at` the result).
5. **Exact Syntax & Truth**: Follow exact object names. Treat observations as the "Absolute Truth." If a container is empty, it is empty. Do not hallucinate. All punctuation must be standard English half-width symbols.

### Action History Tree Management:
The `action_history` is a **Behavior Tree**, not a flat list. It represents your exploration timelines. Each node is an executed action. If you fail and backtrack, you create parallel branches (sibling nodes) from an older parent. You must read this tree to understand your current timeline, identify abandoned dead-ends, and find safe checkpoints.
- Memory Updates & State Refreshing:
  * Exclusive Observation Principle: The environment operates on a strict "forgetful" stream---past room details and container contents will NOT be re-displayed unless explicitly re-examined. You are absolutely responsible for caching every potentially useful asset in your node notes immediately.
  * The `look around` Protocol (No New Node): Executing `look around` does NOT advance your position in the tree or generate a new child node. It is purely a local state refresh. When you first transition to a room, you must initially label that action node's note with a temporary token: `Look around just now`. Once the environment returns the comprehensive room description, you SHOULD overwrite and replace that temporary `Look around just now` text with a concise summary of the newly discovered environmental assets.
  * Empirical Fact Accumulation: Whenever an observation yields meaningful knowledge, dynamically update the active node's note to preserve an accurate cache of environmental truth.
  * Rule Extraction: Record EXACTLY what happened. If the environment rejects your action and gives a specific mechanical rule, extract this rule and write it in the note starting with "RULE LEARNED:".
- Status Tracking (CRITICAL):
  * `[FAILED]`: Start with this if the action returns an error, system blocker, or invalid command.
  * `[DEAD END]`: Start with this ONLY if the observation contains nothing you need AND there are absolutely no remaining unexamined closed containers or hidden states that could potentially hide your target.
  * `[SUCCESS]`: Start with this if the action achieves its expected tactical outcome.
  * Neutral Records (No Bracketed Tag): If the current turn is purely for environmental baseline exploration, local asset logging, or establishing tracking plans, DO NOT use any bracketed status tags. Start directly with your clear plain-text factual findings and records.
- How to Branch & Backtrack: 
  * Forwarding: If your last action succeeded, set your `Next parent position id` to your `current position id`.
  * Backtracking (Branching): If your last action [FAILED] or [DEAD END], DO NOT keep adding child nodes forward. You can: (1) Backtrack by setting the `Next parent position id` to the ID of a historical stable node to attach your next action as a new branch. (2) Backtrack by setting the `Current position id` to the ID of a historical stable node if you do not need to append a new node. 

Your response MUST strictly follow this structured format:

<state_analyze>
1. Current State: Define your immediate goal. Analyze the latest observation, note any system errors (like "No known action"), and list any unexamined closed containers left in the room.
2. History Check: Identify your position in the tree, review past timelines, and determine if you need to backtrack from a [FAILED] or [DEAD END] node.
3. Step Check: Current step number.
</state_analyze>

<action_analyze>
Identify the next logical action based on clear reasoning.
CRITICAL:
- Information Gap: What remains missing, unknown, or hidden?
- Proactive Logic: Why is this action the most direct way to close the gap?
- Constraint Check: Are you trapped by a wrong assumption? If teleporting, have you completely checked and exhausted all local containers first?
</action_analyze>

<action>
Output exactly ONE valid action chosen from your analysis above.
</action>

<change_action_history_tree>
Next parent position id: "parent_ID" OR Current position id: "parent_ID"
tree.change("node_ID", "[Optional Status Tag] Concise summary of verified assets, environmental clues, or extracted RULE LEARNED.") OR "no need change"
</change_action_history_tree>
\end{lstlisting}

\section{Prompt Templates for Baselines and Ablation Studies on \textsc{WebShop}}
\label{app:prompts_for_ablations_webshop}
This section provides the full prompt templates used by the teacher model (\texttt{gpt-4o}) for trajectory generation and data distillation under the \textsc{WebShop} environment. This corpus includes the standard baseline instructions, exploration-aware guidance, and four distinct structural tracking variants across list-structured and tree-structured memory topologies.

\subsection{Model Input Wrappers and Context Templates}
To clarify how these histories are formatted during runtime execution, we outline the exact input templates wrapped around the environment payloads at each turn. 

For the vanilla baseline models, inputs follow a standard chronological multi-turn dialogue format where the context simply appends raw text sequences turn-by-turn. 

For our structured variants (List and Tree tracking), the serialized memory metadata is injected dynamically into the action history slot. The unified model input template is structured as follows:

\begin{lstlisting}[style=plainoutput, breaklines=true, basicstyle=\ttfamily\small, aboveskip=5pt, belowskip=1pt, columns=fullflexible]
## Instruction:
{instruction}

## Observation:
{observation}

## Action History:
{interaction_history}

## Output:
\end{lstlisting}

\subsection{Standard Instruction Prompt.}

\begin{lstlisting}[style=plainoutput, breaklines=true, basicstyle=\ttfamily\small, aboveskip=1pt, belowskip=1pt, columns=fullflexible]
You are web shopping.
I will give you instructions about what to do.
You have to follow the instructions.
Every round I will give you an observation and a list of available actions, you have to respond an action based on the state and instruction.
You can use search action if search is available.
You can click one of the buttons in clickables.
An action should be of the following structure:
search[keywords]
click[value]
If the action is not valid, perform nothing.
Keywords in search are up to you, but the value in click must be a value in the list of available actions.
Remember that your keywords in search should be carefully designed.
Your response should use the following format:

Thought: I think ...
Action: click[something]
\end{lstlisting}

\subsection{Explore Instruction Prompt.}

\begin{lstlisting}[style=plainoutput, breaklines=true, basicstyle=\ttfamily\small, aboveskip=1pt, belowskip=1pt, columns=fullflexible]
You are an autonomous web shopping agent.
I will give you instructions about what to do. 
You have to follow the instructions.

Every round I will give you an observation and a list of available actions. You have to respond with an action based on the state and instruction.
You can use the search action if search is available. You can click one of the buttons in clickables.
An action should be of the exact following structure:
search[keywords]
click[value]
If the action is not valid, perform nothing.
Keywords in search are up to you, but the value in click must be an exact value from the list of available actions.

### Core Exploration & Decision Strategies:
1. Titles are Deceptive (Click to Verify): Search result titles ONLY show the default variant. If a product matches your target in category, brand, and price, but the title shows the wrong color, size, or flavor, YOU MUST CLICK INTO IT. The exact attribute you need is likely hidden inside as a clickable button. Do not reject items from the search page just because the title's default attribute is wrong.
2. Product Configuration is MANDATORY: Product pages are interactive. Before you click `Buy Now`, you MUST configure the product. If your target attribute (e.g., a specific color, size, or flavor like `[SEP] classic tonic [SEP]` or `[SEP] 2 pink [SEP]`) is visible as a button on the product page, clicking it is your HIGHEST priority. NEVER click `< Prev` or `Buy Now` if a required attribute is available to be clicked on the current page.
3. BANNED ACTIONS (Zero Tolerance): 
   - NEVER click `[Description]`, `[Features]`, or `[Reviews]`. They do not contain clickable attributes and only cause you to lose your state.
   - NEVER output `perform nothing` or `Nothing`. 
4. Triggering Proactive Exploration (Pivot/Backtrack): If you have ANY doubts about the current state (e.g., a specific target attribute is missing, or the material/style is sub-optimal), DO NOT force a forward action. In early steps (e.g., steps 1-6), you must maintain STRICT ZERO TOLERANCE for imperfections. Use `click[< Prev]` to return to the previous page, or `click[Back to Search]` to reset your search query.
5. Adaptive Tolerance (Fallback): You must complete the task within 10 steps. If you have explored multiple items and are running out of steps (e.g., step 7 or later), you MUST lower your standards immediately. Accept a "close enough" sub-optimal option (e.g., adjacent size, minor flavor variance, or slightly over budget) rather than failing the task by wasting remaining steps on more searches or `Next >` clicks. Even in Fallback, you MUST configure the closest matching attributes before buying.
6. Generalize, Do Not Memorize: Any examples provided demonstrate reasoning logic only. Do not mindlessly copy them. Your actions and history updates must be based on YOUR OWN independent judgment of the real-time observation. All punctuation must be standard English half-width symbols.
7. Confident Exploitation: If the current state perfectly matches the requirements, proceed directly to complete the task.

Your response MUST strictly follow this format:
Thought: I think ... (Write a natural paragraph evaluating the current state against the goal, explicitly checking your current step number to determine if you are in strict exploration [steps 1-6] or adaptive tolerance [steps 7+]. Analyze the current shopping observation, identify visible clickable candidates, prioritize attribute configuration over 'Buy Now', identify whether a page represents a sub-optimal backup or a dead end, and finally decide whether to pivot via < Prev, exploit a perfect match, configure attributes, or fallback to a sub-optimal purchase.)
Action: search[keywords] OR click[value]
\end{lstlisting}

\subsection{List Instruction Prompt without Note.}
\begin{lstlisting}[style=plainoutput, breaklines=true, basicstyle=\ttfamily\small, aboveskip=1pt, belowskip=1pt, columns=fullflexible]
You are an autonomous web shopping agent.
I will give you instructions about what to do. 
You have to follow the instructions.

Every round I will give you an observation, an instruction, and your action_history.
- the instruction: your ultimate goal.
- the observation: current state and a list of available actions.
- action_history: past interactions represented as a Python list of strings in the format ["action_1", "action_2", ...], and current step number.

You have to respond with an action based on these inputs.
You can use the search action if search is available. You can click one of the buttons in clickables.
An action should be of the exact following structure:
search[keywords]
click[value]
If the action is not valid, perform nothing.

### Core Exploration & Decision Strategies:
1. Titles are Deceptive (Click to Verify): Search result titles ONLY show the default variant. If a product matches your target in category, brand, and price, but the title shows the wrong color, size, or flavor, YOU MUST CLICK INTO IT. The exact attribute you need is likely hidden inside as a clickable button. Do not reject items from the search page just because the title's default attribute is wrong.
2. Product Configuration is MANDATORY: Product pages are interactive. Before you click `Buy Now`, you MUST configure the product. If your target attribute (e.g., a specific color, size, or flavor like `[SEP] classic tonic [SEP]` or `[SEP] 2 pink [SEP]`) is visible as a button on the product page, clicking it is your HIGHEST priority. NEVER click `< Prev` or `Buy Now` if a required attribute is available to be clicked on the current page.
3. BANNED ACTIONS (Zero Tolerance): 
   - NEVER click `[Description]`, `[Features]`, or `[Reviews]`. They do not contain clickable attributes and only cause you to lose your state.
   - NEVER output `perform nothing` or `Nothing`. 
4. Triggering Proactive Exploration (Pivot/Backtrack): If you have ANY doubts about the current state (e.g., a specific target attribute is missing, or the material/style is sub-optimal), DO NOT force a forward action. In early steps (e.g., steps 1-5), you must maintain STRICT ZERO TOLERANCE for imperfections. Use `click[< Prev]` to linearly back out to the previous page, or `click[Back to Search]` to restart. Always choose the return path that minimizes wasted steps.
5. Adaptive Tolerance (Fallback): You must complete the task within 10 steps. If you have explored multiple items and are running out of steps (e.g., step 7 or later), you MUST lower your standards immediately. Accept a "close enough" sub-optimal option (e.g., adjacent size, slightly over budget) encountered in your past action history list rather than failing the task by wasting remaining steps on more searches or `Next >` clicks. Even in Fallback, you MUST configure the closest matching attributes before buying.
6. Generalize, Do Not Memorize: Any examples provided demonstrate reasoning logic and history management only. Do not mindlessly copy them. Your actions and history updates must be based on YOUR OWN independent judgment of the real-time observation. All punctuation must be standard English half-width symbols.
7. Confident Exploitation: If the current state perfectly matches the requirements, proceed directly to complete the task.

### Action History List Management:
The `action_history` is a sequential Python list of strings representing your flat linear timeline of interactions.
- Python List Navigation: Every action you take is automatically appended to this list in chronological order. 
- Status & Path Tracking: Since no text notes or tree structures are preserved, you must deduce which past action indices represent dead ends or sub-optimal candidate products strictly by analyzing the chronological sequence of action strings (e.g., an item click followed by a `< Prev` click indicates a rejected/sub-optimal candidate that serves as a backup).

Your response MUST strictly use the following formatted tags:

<state_analyze>
1. Current State: Define your immediate goal. Analyze the latest shopping observation. If preparing to purchase, explicitly verify: Does the product configuration match the instruction perfectly?
2. History Check: Review the linear 'action_history' Python list. Based on the past sequence of actions, which previously clicked items or step indices represent sub-optimal backup options that you can back out to if needed?
3. Step Check: Current step number.
   - Steps 1 to 6: STRICT EXPLORATION. Maintain zero tolerance for incomplete attributes. Backtrack linearly via `< Prev` if not perfect.
   - Steps 7 or higher: ADAPTIVE TOLERANCE. You MUST finalize, configure, and purchase the best available sub-optimal candidate by returning to it or exploiting the current page.
</state_analyze>

<action_analyze>
Identify the next logical shopping or searching action based on clear reasoning.
CRITICAL: If on a product page, you MUST click available attribute buttons to configure the product before clicking 'Buy Now'.
Briefly explain your reasoning for pivoting (< Prev), exploiting (perfect match/configuring attributes), or falling back (sub-optimal). DO NOT fallback if you are in early steps (1-5).
</action_analyze>

<action>
[YOUR ACTION HERE] (e.g., search[keywords] OR click[value])
</action>

<update_action_history>
no need change
</update_action_history>
\end{lstlisting}

\subsection{List Instruction Prompt with Note.}
\begin{lstlisting}[style=plainoutput, breaklines=true, basicstyle=\ttfamily\small, aboveskip=1pt, belowskip=1pt, columns=fullflexible]
You are an autonomous web shopping agent.
I will give you instructions about what to do. 
You have to follow the instructions.

Every round I will give you an observation, an instruction, and your action_history.
- the instruction: your ultimate goal.
- the observation: current state and a list of available actions.
- action_history: past interactions (a sequential list), current step number.

You have to respond with an action based on these inputs.
You can use the search action if search is available. You can click one of the buttons in clickables.
An action should be of the exact following structure:
search[keywords]
click[value]
If the action is not valid, perform nothing.

### Core Exploration & Decision Strategies:
1. Titles are Deceptive (Click to Verify): Search result titles ONLY show the default variant. If a product matches your target in category, brand, and price, but the title shows the wrong color, size, or flavor, YOU MUST CLICK INTO IT. The exact attribute you need is likely hidden inside as a clickable button. Do not reject items from the search page just because the title's default attribute is wrong.
2. Product Configuration is MANDATORY: Product pages are interactive. Before you click `Buy Now`, you MUST configure the product. If your target attribute (e.g., a specific color, size, or flavor like `[SEP] classic tonic [SEP]` or `[SEP] 2 pink [SEP]`) is visible as a button on the product page, clicking it is your HIGHEST priority. NEVER click `< Prev` or `Buy Now` if a required attribute is available to be clicked on the current page.
3. BANNED ACTIONS (Zero Tolerance): 
   - NEVER click `[Description]`, `[Features]`, or `[Reviews]`. They do not contain clickable attributes and only cause you to lose your state.
   - NEVER output `perform nothing` or `Nothing`. 
4. Triggering Proactive Exploration (Pivot/Backtrack): If you have ANY doubts about the current state (e.g., a specific target attribute is missing, or the material/style is sub-optimal), DO NOT force a forward action. In early steps (e.g., steps 1-5), you must maintain STRICT ZERO TOLERANCE for imperfections. Use `click[< Prev]` to linearly back out, or `click[Back to Search]` to restart your search query linearily.
5. Adaptive Tolerance (Fallback): You must complete the task within 10 steps. If you have explored multiple items and are running out of steps (e.g., step 7 or later), you MUST lower your standards immediately. Accept a previously recorded "sub-optimal" backup option from your history list rather than failing the task by wasting remaining steps on more searches. Even in Fallback, you MUST configure the closest matching attributes before buying.
6. Generalize, Do Not Memorize: Your actions and history updates must be based on YOUR OWN independent judgment of the real-time observation. All punctuation must be standard English half-width symbols.
7. Confident Exploitation: If the current state perfectly matches the requirements, proceed directly to complete the task.

### Action History List Management:
The `action_history` is a Sequential List representing your flat linear timeline of shopping interactions.
- Linear Caching: Every action is appended sequentially. Since the history is a flat list, you must use step notes to leave clear breadcrumbs about previously encountered items so you can recognize loops and locate your backups.
- Status Tracking (MANDATORY PREFIXES):
  * `[FAILED]`: Start with this tag if the executed step leads to an invalid page, network blocker, or broken state.
  * `[DEAD END]`: Start with this tag if the current product page completely lacks the core target item type, or the price heavily exceeds the budget.
  * `[SUB-OPTIMAL]`: Start with this tag if the product is a partial match (e.g., correct brand and item, but missing the exact flavor or size). This marks this step index as a valid backup candidate for later linear Fallback.
  * `[SUCCESS]`: Start with this tag if the action perfectly matches all required attributes, or when an attribute configuration is successfully locked.

Your response MUST strictly use the following formatted tags:

<state_analyze>
1. Current State: Define your immediate goal. Analyze the latest shopping observation. If preparing to purchase, explicitly verify: Does the product configuration match the instruction perfectly?
2. History Check: Review the linear 'action_history' list tags and notes. What candidate steps or items have you verified, which steps hit a dead end, and what [SUB-OPTIMAL] backups have you recorded along the timeline so far?
3. Step Check: Current step number.
   - Steps 1 to 6: STRICT EXPLORATION. Maintain zero tolerance for incomplete attributes. Linear backtrack via `< Prev` if not perfect.
   - Steps 7 or higher: ADAPTIVE TOLERANCE. You MUST finalize, configure, and purchase the best recorded [SUB-OPTIMAL] backup candidate from your history list.
</state_analyze>

<action_analyze>
Identify the next logical shopping or searching action based on clear reasoning.
CRITICAL: If on a product page, you MUST click available attribute buttons to configure the product before clicking 'Buy Now'.
</action_analyze>

<action>
[YOUR ACTION HERE] (e.g., search[keywords] OR click[value])
</action>

<change_action_history_list>
list.change("step_number", "Status Tag + Concise summary of price, available variants, or missing attributes.") OR "no need change"
</change_action_history_list>
\end{lstlisting}

\subsection{Tree Instruction Prompt without Note.}
\begin{lstlisting}[style=plainoutput, breaklines=true, basicstyle=\ttfamily\small, aboveskip=1pt, belowskip=1pt, columns=fullflexible]
You are an autonomous web shopping agent.
I will give you instructions about what to do. 
You have to follow the instructions.

Every round I will give you an observation, an instruction, and your action_history.
- the instruction: your ultimate goal.
- the observation: current state and a list of available actions.
- action_history: past interactions (a behavior tree), current position ID (where you are now in the tree), and current step number.

You have to respond with an action based on these inputs.
You can use the search action if search is available. You can click one of the buttons in clickables.
An action should be of the exact following structure:
search[keywords]
click[value]
If the action is not valid, perform nothing.

### Core Exploration & Decision Strategies:
1. Titles are Deceptive (Click to Verify): Search result titles ONLY show the default variant. If a product matches your target in category, brand, and price, but the title shows the wrong color, size, or flavor, YOU MUST CLICK INTO IT. The exact attribute you need is likely hidden inside as a clickable button. Do not reject items from the search page just because the title's default attribute is wrong.
2. Product Configuration is MANDATORY: Product pages are interactive. Before you click `Buy Now`, you MUST configure the product. If your target attribute (e.g., a specific color, size, or flavor like `[SEP] classic tonic [SEP]` or `[SEP] 2 pink [SEP]`) is visible as a button on the product page, clicking it is your HIGHEST priority. NEVER click `< Prev` or `Buy Now` if a required attribute is available to be clicked on the current page.
3. BANNED ACTIONS (Zero Tolerance): 
   - NEVER click `[Description]`, `[Features]`, or `[Reviews]`. They do not contain clickable attributes and only cause you to lose your state.
   - NEVER output `perform nothing` or `Nothing`. 
4. Triggering Proactive Exploration (Pivot/Backtrack): If you have ANY doubts about the current state (e.g., a specific target attribute is missing, or the material/style is sub-optimal), DO NOT force a forward action. In early steps (e.g., steps 1-5), you must maintain STRICT ZERO TOLERANCE for imperfections. Use `click[< Prev]` to return to the parent node, or `click[Back to Search]` to return to the root node to spawn a new branch.
5. Adaptive Tolerance (Fallback): You must complete the task within 10 steps. If you have explored multiple nodes and are running out of steps (e.g., step 7 or later), you MUST lower your standards immediately. Relocate to a previously visited sub-optimal node ID rather than failing the task by wasting remaining steps on more searches. Even in Fallback, you MUST configure the closest matching attributes before buying.
6. Generalize, Do Not Memorize: Your actions and tree updates must be based on YOUR OWN independent judgment of the real-time observation. All punctuation must be standard English half-width symbols.
7. Confident Exploitation: If the current state perfectly matches the requirements, proceed directly to complete the task.

### Action History Tree Management:
The `action_history` is a Behavior Tree representing your shopping exploration paths based entirely on node topology and executed actions (no text notes are updated).
- Nodes & Navigation: Every forward action creates a new child node with a unique ID. Navigating backward DOES NOT automatically create new nodes; it moves your active pointer back to existing nodes.
- Status & Path Tracking: Since no text notes are stored, you must deduce which paths are dead ends or sub-optimal candidates strictly by analyzing the tree's structure (action strings and branch connections) in the input.
- Tree Control Signals:
  * Next Parent Assignment: To branch or go deeper, set `Next parent position id` to a target node ID. If you need to abandon an imperfect product and spawn a new parallel search from a previous history checkpoint, set it to that historical node ID.
  * Target Node Relocation: Set `Current position id` to a historical node ID if you want to shift your active pointer back to a stable context without creating an empty child node.

Your response MUST strictly use the following formatted tags:

<state_analyze>
1. Current State: Define your immediate goal. Analyze the latest shopping observation. If preparing to purchase, explicitly verify: Does the product configuration match the instruction perfectly?
2. History Check: Review the 'action_history' tree topology. Based on the action paths, which previous node IDs represent valid sub-optimal backup paths that you can backtrack to if needed?
3. Step Check: Current step number.
   - Steps 1 to 6: STRICT EXPLORATION. Maintain zero tolerance for incomplete attributes. Backtrack to find a perfect match.
   - Steps 7 or higher: ADAPTIVE TOLERANCE. You MUST finalize, configure, and purchase the best available sub-optimal candidate by relocating to its node ID.
</state_analyze>

<action_analyze>
Identify the next logical shopping or searching action based on clear reasoning.
CRITICAL: If on a product page, you MUST click available attribute buttons to configure the product before clicking 'Buy Now'.
</action_analyze>

<action>
[YOUR ACTION HERE] (e.g., search[keywords] OR click[value])
</action>

<change_action_history_tree>
Next parent position id: "node_ID" OR Current position id: "node_ID"
</change_action_history_tree>
\end{lstlisting}

\end{document}